%% file: main.tex
\documentclass{article}

\usepackage{microtype}
\usepackage{graphicx}
\usepackage{subfigure}
\usepackage{booktabs} 

\usepackage{hyperref}

\usepackage[accepted]{icml2024}

\usepackage{amsmath}
\usepackage{amssymb}
\usepackage{mathtools}
\usepackage{amsthm}

\usepackage[capitalize,noabbrev]{cleveref}

\theoremstyle{plain}
\newtheorem{theorem}{Theorem}[section]
\newtheorem{proposition}[theorem]{Proposition}

\newtheorem{corollary}[theorem]{Corollary}
\theoremstyle{definition}
\newtheorem{definition}[theorem]{Definition}

\theoremstyle{remark}
\newtheorem{remark}[theorem]{Remark}

\usepackage[textsize=tiny]{todonotes}

\input{math_commands.tex}

\usepackage{comment}
\usepackage{adjustbox}
\usepackage{tabularx,multirow,longtable}

\usepackage{pifont}
\newcommand{\cmark}{\ding{51}}%
\newcommand{\xmark}{\ding{55}}%
\usepackage{wrapfig}
\usepackage{fontawesome}

\icmltitlerunning{Multi-objective Bayesian optimization with tied multivariate ranks}

\begin{document}

\twocolumn[
\icmltitle{BOtied: Multi-objective Bayesian optimization with tied multivariate ranks}



\icmlsetsymbol{equal}{*}

\begin{icmlauthorlist}
\icmlauthor{Ji~Won Park}{equal,prescient}
\icmlauthor{Nata\v{s}a Tagasovska}{equal,prescient}
\icmlauthor{Michael Maser}{prescient}
\icmlauthor{Stephen Ra}{prescient}
\icmlauthor{Kyunghyun Cho}{prescient,comp,cds}
\end{icmlauthorlist}

\icmlaffiliation{prescient}{Prescient Design, Genentech, South San Francisco, USA}
\icmlaffiliation{comp}{Department of Computer Science, New York University, New York City, USA}
\icmlaffiliation{cds}{Center for Data Science, New York University, New York City, USA}

\icmlcorrespondingauthor{Ji~Won Park}{park.ji\_won@gene.com}
\icmlcorrespondingauthor{Nata\v{s}a Tagasovska}{natasa.tagasovska@roche.com}

\icmlkeywords{Bayesian optimization, multi-objective optimization, density estimation, copulas}

\vskip 0.3in
]



\printAffiliationsAndNotice{\icmlEqualContribution} 

\begin{abstract}
Many scientific and industrial applications require the joint optimization of multiple, potentially competing objectives. Multi-objective Bayesian optimization (MOBO) is a sample-efficient framework for identifying Pareto-optimal solutions. At the heart of MOBO is the acquisition function, which determines the next candidate to evaluate by navigating the best compromises among the objectives. In this paper, we show a natural connection between non-dominated solutions and the extreme quantile of the joint cumulative distribution function (CDF). Motivated by this link, we propose the Pareto-compliant CDF indicator and the associated acquisition function, BOtied. BOtied inherits desirable invariance properties of the CDF, and an efficient implementation with copulas allows it to scale to many objectives. Our experiments on a variety of synthetic and real-world problems demonstrate that BOtied outperforms state-of-the-art MOBO acquisition functions while being computationally efficient for many objectives. 
\end{abstract}

\section{Introduction}
\label{sec:intro}

\begin{figure}[t!]
    \centering
    \includegraphics[trim=0cm 0.05cm 0cm 0cm, clip, width=0.49\textwidth]{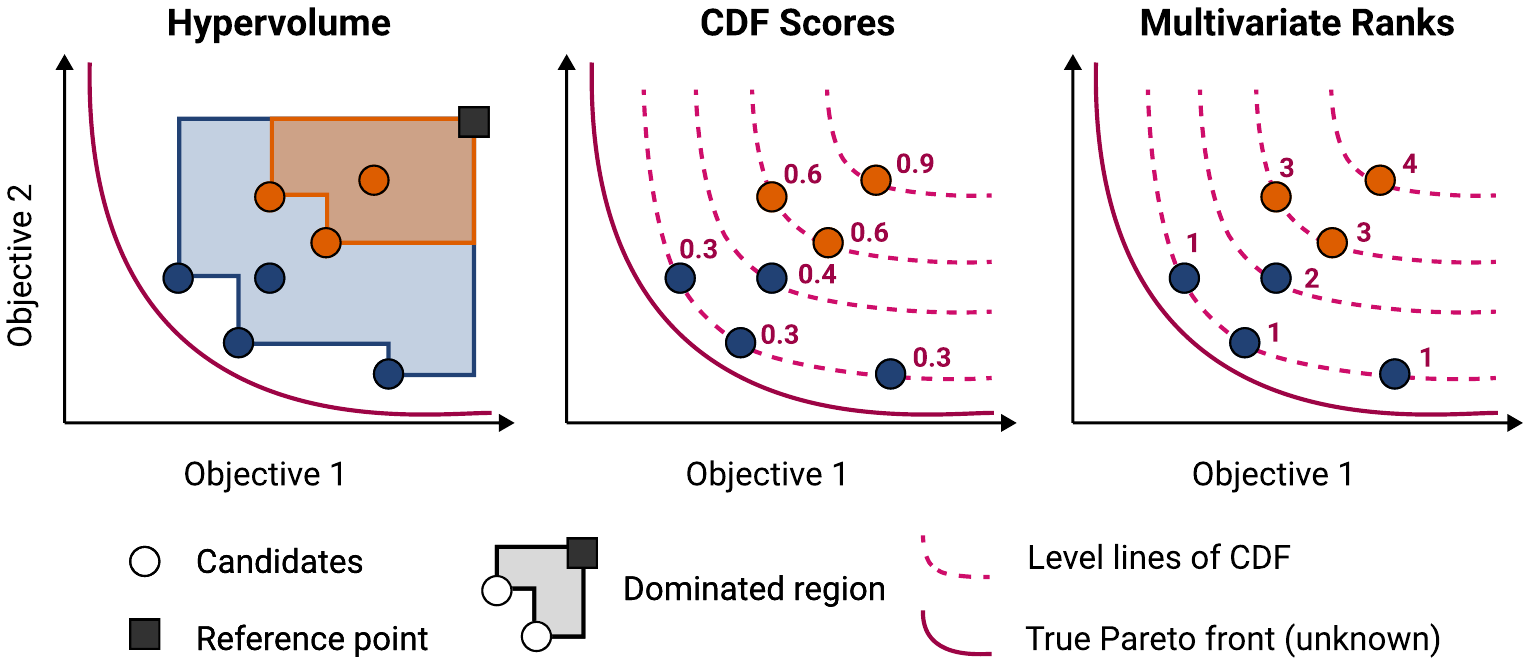}
    \vspace{-0.5cm}
     \caption{Illustration of the conceptual link between the empirical Pareto front probed by the HV indicator and innermost level line of the CDF probed by the {BOtied} CDF indicator. The blue set of candidates dominates the orange. The HV indicator is consistent with this ordering; the area of the box dominated by the blue set is greater. The {BOtied} CDF values and associated multivariate ranks also favor the blue.}  
     \label{fig:intro}
     \vspace{-0.5cm}
 \end{figure}
 
Bayesian optimization (BO) has demonstrated promise in a variety of scientific and industrial domains where the goal is to optimize an expensive black-box function using a limited number of potentially noisy function evaluations \citep{romero2013navigating,calandra2016bayesian,kusne2020fly,shields2021bayesian,zuo2021accelerating,bellamy2022batched,park2022propertydag}. In BO, we fit a probabilistic surrogate model on the available observations so far. Based on the model, the acquisition function determines the next candidate to evaluate by balancing exploration (evaluating highly uncertain candidates) with exploitation (evaluating designs believed to be optimal).
Often, applications call for the joint optimization of $M{\rm \geq}2$ multiple, potentially competing objectives \citep{marler2004survey, jain2017biophysical, tagasovska2022pareto}. Unlike in single-objective settings, a single optimal solution may not exist and we must identify a set of solutions that represents the best compromises among the multiple objectives. The acquisition function in multi-objective Bayesian optimization (MOBO) navigates these trade-offs as it guides the optimization toward regions of interest. 

Many bona fide MO acquisition functions without scalarization
involve high-dimensional integrals and scale poorly with increasing $M$. Moreover, improvement-based acquisition functions including some variants of random scalarization are sensitive to non-informative monotonic transformations of the objectives. This is a pain point for many practical applications. For instance, in biochemistry, the dissociation constant $K_D$ is typically expressed in terms of its log transformation, the ${\rm p}K_D \equiv {\rm -}\log_{10}(K_D)$. It would be desirable to work with acquisition functions that are invariant to the choice of such unit conversions.


To address these challenges, 
we propose {BOtied}\footnote{The name stems from non-dominated solutions considered to be "tied."}, a novel acquisition function based on multivariate ranks. We show that {BOtied} has the desirable property of being invariant to relative rescaling or monotonic transformations of the objectives. While it maintains the multivariate structure of the objective space, its implementation has highly favorable time complexity and we report wall-clock time competitive with random scalarization.

In \autoref{fig:intro}, we present the intuition behind {BOtied}.  
Consider a minimization setup for $M{\rm=}2$ where we seek to identify solutions on the true Pareto front (solid red curves), unknown to us. 
Suppose we have many candidates, represented as circular posterior blobs in the objective space, where the predictive distributions have been output by our surrogate model. For simplicity, assume the uncertainties (sizes of blobs) are comparable among the candidates.
How do we estimate the quality of our solutions (i.e., each candidate set's proximity to the true Pareto front)? 
One quality indicator is the hypervolume \citep[HV;][]{zitzler2003performance}, defined as the size of the polytope bounded from above by a predefined reference point and dominated by the candidate set (shaded areas in the leftmost panel). When one candidate set dominates another, its HV is greater. We can visually confirm that the HV of the blue Pareto approximation is greater than that of the orange. 

\begin{figure}[t!]
    \centering
    \includegraphics[trim=0cm 0.1cm 0cm 0cm, clip, width=0.49\textwidth]{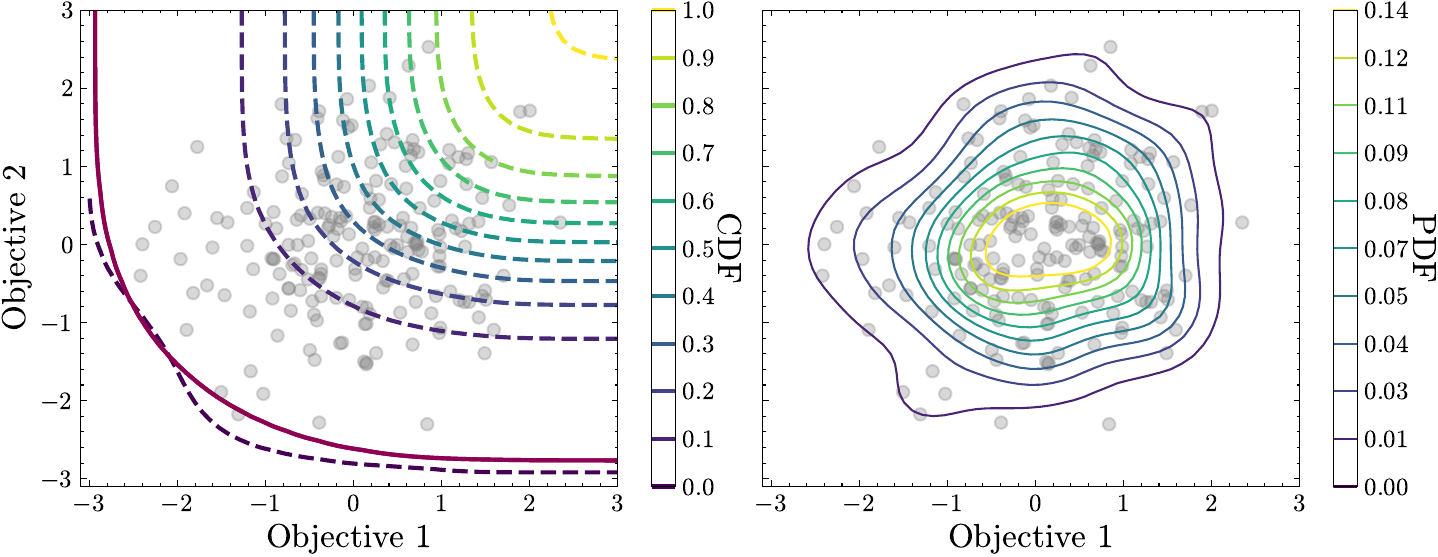}
    \vspace{-0.5cm}
     \caption{Level lines of the CDF (left) and the PDF (right) from kernel density estimation based on 200 observations (gray dots). The zero level line of the CDF closely traces the true Pareto front (solid red curve).}
     \label{fig:intro_fig2}
     \vspace{-0.5cm}
 \end{figure}

Next, let us adopt the related, but distinct, perspective of multivariate ranking \citep[see, e.g.,][]{ghosal2022multivariate}. Define the random objective vector $Y = [f_1(X), \dotsc, f_M(X)]$ taking values in $\mathbb{R}^M$, which is the result of applying the objective function $f: \mathbb{R}^d \rightarrow \mathbb{R}^M$ on the random design variable $X$ that takes values in $\mathbb{R}^d$.  
How do we compare two realizations of $Y$, say ${\bm y}$ and ${\bm y'}$?

Ranking vectors is non-trivial, as there is no natural ordering in Euclidean spaces when $M{\rm \geq}2$. We propose to use the joint cumulative distribution function (CDF), defined as the probability of ${\bm y}$ being weakly dominated: $F_Y({\bm y}) = \mathbb{P}(Y_1{\rm  \leq}y_1, \dotsc, Y_M{\rm \leq}y_M)$, where ${\bm y}= [{ y}_1, \dotsc, {y}_M] \in \mathbb{R}^M$. The CDF formalizes the rank ordering of vectors as weak dominance in the joint minimization of $M$ objectives \citep{binois2015estimation}. Specifically, The CDF scores and their $\alpha$-level lines $L_\alpha^{F_Y} = \{{\bm y}: F_Y({\bm y}) = \alpha \}$ are depicted in the middle panel of \autoref{fig:intro} for multiple values of $\alpha \in [0, 1]$. All candidates with equal multivariate rank, or ties, lie on the same level line, as shown in the rightmost panel.


Multivariate ranking via the CDF can be understood in relation to the associated probability density function (PDF), as the CDF is the integral of the PDF. \autoref{fig:intro_fig2} shows the CDF and the associated PDF side by side for a bi-objective setting ($M{\rm =}2$). 
The right panel shows the PDF fit on 200 outcome samples (gray dots) via kernel density estimation, where the outcome samples were drawn from an elliptical Gaussian.
The left panel shows the level lines of the corresponding CDF. The $\alpha$-level lines converge to the approximate Pareto front as $\alpha \rightarrow 0$. The lowermost level line ($\alpha{\rm\approx}0$) closely traces the convex shape of the true Pareto front shown as the solid red curve.

\paragraph{Contributions} Motivated by the interpretation of multivariate ranks as a MO indicator, we make the following contributions:
(i) We propose a new Pareto-compliant performance criterion, the {CDF} indicator  (\autoref{sec:background});
(ii) We propose a scalable and robust acquisition function based on the CDF and associated multivariate ranks, which we call {BOtied} (\autoref{sec:method}); and
(iii) We release the full codebase implementing our evaluations of MOBO acquisitions in a variety of synthetic and real-world data scenarios (\autoref{sec:experiments}).

\section{Related work}

\textbf{MO indicators and acquisition functions.} A computationally attractive approach to MOBO scalarizes the objectives with random preference weights \citep{knowles2006parego,paria2020flexible} and applies a single-objective acquisition function. The distribution of the weights, however, may be insufficient to encourage exploration when there are many objectives with unknown scales. 

Alternatively, we may preserve the MO structure by seeking improvement on a set-based performance metric, such as the HV indicator \citep{embrechts2003using} or the R2 indicator \citep{deutz2019expected,deutz2019r2}. Improvement-based acquisition functions such as the expected hypervolume improvement \citep[EHVI;][]{emmerich2011hypervolume,daulton2020differentiable,daulton2021parallel} are sensitive to the rescaling of the objectives, which may carry drastically different natural units. 
In particular, computing the HV has time complexity that is super-polynomial in the number of objectives, because it entails computing the volume of an irregular polytope \citep{yang2019efficient}. Despite the efficiency improvement achieved by box decomposition algorithms \citep{dachert2017efficient,yang2019efficient}, HV computation remains slow when $M{\rm >}4$. 

Another class of acquisition strategies is entropy search, which focuses on maximizing the information gain from the next observation \citep{villemonteix2009informational,hennig2012entropy,hoffman2015output,shah2015parallel,hernandez2016general,belakaria2019max,tu2022joint}. Entropy searches are commonly implemented in box decompositions as well, but are costly to evaluate without using more tractable bounds to serve as approximations.
 


\textbf{Multivariate ranking.} 
The scale-invariant properties of ranking makes it an attractive tool for optimization. \citet{binois2015estimation} relates the Pareto front to the extreme level line of the CDF, $F_Y$. Considering ranking dominance as an alternative to Pareto dominance, \citet{kukkonen2007ranking} propose computing the ranks of individual objectives separately and combining them post-hoc with a simple aggregation function (min, max, average) to obtain the overall fitness value for a given candidate. \citet{binois2020kalai} explores the question of how to choose from the set of non-dominated solutions, which grows with $M$, and makes a game-theoretic argument for how to make the compromise. In particular, they define trade-offs in the copula space, which is the scale-invariant rank transformation of the original objective function. For single-objective BO, \citet{picheny2019ordinal} propose Ordinal BO, which uses a Gaussian process (GP) surrogate that is only sensitive to the rankings of the inputs and objective values. Notably, this method can robustly handle ill-conditioned and multi-modal distributions in the objective function values, for which GP models are known to often fail. Similarly, \citet{eriksson2021scalable} use the rank transformations to magnify values at the end of the observed ranges. To our knowledge, however, our work is the first to incorporate multivariate rankings enabled by the \textit{joint} CDF into a MOBO algorithm. We explicitly account for the structure of the $M$-variate objective distribution in identifying the full Pareto front.  

A more detailed overview and positioning of {BOtied} with respect to the MO literature can be found in \autoref{app:related_work}.

\section{Background}
\label{sec:background}

\subsection{Bayesian Optimization}

Bayesian optimization (BO) is a popular technique for sample-efficient black-box optimization \citep[see][for a review]{shahriari2015taking, frazier2018tutorial}. 
In a single-objective setting, suppose our objective $f: \mathcal{X} \rightarrow \mathbb{R}$ is a black-box function of the design space $\mathcal{X}$ that is expensive to evaluate. 
Our goal is to efficiently identify a design $\bm{x}^\star \in \mathcal{X}$ minimizing\footnote{
    For simplicity, we define the task as minimization in this paper without loss of generality. 
    For maximization, we can negate $f$, for instance.
} 
$f$. 
BO leverages two tools, a probabilistic surrogate model and a utility function, to trade off exploration (evaluating highly uncertain designs) and exploitation (evaluating designs believed to minimize $f$) in a principled manner. 

For each iteration $t$, we have a dataset $\mathcal{D}_t = \{( \bm{x}^{(1)}, {y}^{(1)} ), \cdots, ( \bm{x}^{(N_t)}, {y}^{(N_t)} )\}$, where for each $n \in [N_t]$, ${y}^{(n)}$ is a potentially noisy observation of $f(\bm{x}^{(n)})$.
We first infer the posterior distribution $p({\hat f} | \mathcal{D}_t)$, which serves as a cheap approximation of $f$. 
Next, we introduce a utility function $u: \mathcal{X} \times \mathcal{F} \times \mathscr{D}_t : \rightarrow \mathbb{R}$.
The acquisition function $a(\bm x)$ is simply the expected utility of $\bm x$ with respect to our current belief about $f$:
\begin{align}
    a(\bm x) = \int u(\bm x, \hat f, \mathcal{D}_t) \ p(\hat f | \mathcal{D}_t) \ d\hat f.
\end{align}
For example, we obtain the expected improvement (EI) acquisition function if we take $ u_{\mathrm{EI}}(\bm x, \hat f, \mathcal{D}) = [\hat f(\bm x) - \max_{(\bm x', y') \in \mathcal{D}_t} y']_+,$ where $[\cdot]_+ = \max(\cdot, 0)$ \citep{movckus1975bayesian,jones1998efficient}. 
We select a maximizer of $a$ as the new design, evaluate $f$, and append the observation to the dataset. The surrogate is then refit on the expanded dataset and the procedure repeats. 

\subsection{Multi-objective optimization} \label{sec:moo}

When there are multiple objectives of interest, a single best design may not exist.
Suppose there are $M$ objectives, $f: \mathcal{X} \rightarrow \mathbb{R}^M$.
The goal of MOBO is to identify the set of \textit{Pareto-optimal} solutions such that improving one objective within the set leads to worsening another. 
We say that $\bm x$ dominates $\bm x'$, or ${f}(\bm{x}) \prec {f}(\bm{x}')$, if $f_m(\bm{x}) \leq f_m(\bm{x}')$ for all $m \in [M]$ and $f_m(\bm x) < f_m(\bm x')$ for some $m$.
The set of \textit{non-dominated} solutions $\mathscr{X}^*$ is defined in terms of the Pareto front $\mathcal{P}^*$:
\begin{align} \label{eq:pareto}
\mathscr{X}^\star &= \{\bm{x}: f(\bm{x}) \in \mathcal{P}^\star\}, \nonumber  \\ 
\text{where } \mathcal{P}^\star = \{f(\bm{x}) \: : \: & \bm x \in \mathcal{X}, \;  \nexists \:  \bm{x}' \in \mathcal{X} \textit{ s.t. } f(\bm{x}') \prec f(\bm{x}) \}. \nonumber
\end{align} 
MOBO algorithms typically aim to identify a finite subset of $\mathscr{X}^\star$, which may be infinite, within a given budget of function evaluations. 

\paragraph{Hypervolume}
One way to measure the quality of an approximate Pareto front $\mathcal{P}$ is to compute the hypervolume (HV) ${\rm HV}(\mathcal{P} | \bm{r}_{\rm ref})$ of the polytope bounded from above by $\mathcal{P}$ and from below by $\bm r_{\mathrm{ref}}$, where $\bm r_{\mathrm{ref}} \in \mathbb{R}^M$ is a user-specified \textit{reference point}. More specifically, the HV indicator for a set $A$ is
\begin{align}
    I_{\rm HV}(A) = \int_{\mathbb{R}^M} \mathbb{I}[A \preceq {\bm y} \preceq {\bm r}_{\rm ref}] d{\bm y}.
\end{align}
We obtain the expected hypervolume improvement (EHVI) acquisition function if we take
\begin{align} 
    u_{\mathrm{EHVI}}(\bm x, \hat f, \mathcal{D}) = {\rm HVI}(\mathcal{P}', \mathcal{P} | \bm{r}_{\rm ref}) &=  \nonumber \\ [I_{\rm HV}(\mathcal{P}' | \bm{r}_{\rm ref}) - I_{\rm HV}(\mathcal{P} | \bm{r}_{\rm ref})]_+, \label{eq:ehvi}
\end{align}
where $\mathcal{P}' = \mathcal{P} \cup \{\hat f(\bm x)\}$ \citep{emmerich2005single,emmerich2011hypervolume}.

 

\paragraph{Noisy observations}

In the noiseless setting, the observed baseline Pareto front is the true baseline Pareto front, i.e. $\mathcal{P}_t = \{\bm{y}: \bm{y} \in \mathcal{Y}_t, \: \nexists \: \bm{y}' \in \mathcal{Y}_t \textit{ s.t. } \bm{y}' \prec \bm{y} \}$, where $\mathcal{Y}_t \coloneqq \{\bm{y}^{(n)}\}_{n=1}^{N_t}$. This does not, however, hold in many practical applications, where measurements carry noise. For instance, given a zero-mean Gaussian measurement process with noise covariance $\Sigma$, the feedback for a candidate $\bm{x}$ is $\bm{y} \sim \mathcal{N}\left( {f}(\bm{x}), \Sigma \right)$, not $f(\bm{x})$ itself. 
The \textit{noisy} expected hypervolume improvement (NEHVI) acquisition function marginalizes over the surrogate posterior at the previously observed points $\mathcal{X}_t = \{\bm{x}^{(n)}\}_{n=1}^{N_t}$,
\begin{align} 
    u_{\mathrm{NEHVI}}(\bm x, \hat f, \mathcal{D}) = {\rm HVI}(\hat{\mathcal{P}}_t', \hat{\mathcal{P}}_t | \bm{r}_{\rm ref}) \label{eq:nehvi},
\end{align}
where $\hat{\mathcal{P}}_t = \{\hat f(\bm{x}): \bm x \in \mathcal{X}_t, \;  \nexists \: \bm{x}' \in \mathcal{X}_t \textit{ s.t. } \hat f(\bm{x}') \prec \hat f(\bm{x}) \}$ and $\hat{\mathcal{P}}' = \hat{\mathcal{P}} \cup \{\hat{f}(\bm{x})\}$ \citep{daulton2021parallel}.
\section{Multi-objective BO with multivariate ranks}
\label{sec:method}
In MOBO, it is common to evaluate the quality of an approximate Pareto set ${\mathcal X}$ by computing its distance from the optimal Pareto set $\mathcal{X}^*$ in the objective space, defined by some distance metric $d: 2^{\mathcal{Y}} \times 2^{\mathcal{Y}} \rightarrow \mathbb{R}$ where $2^{\mathcal{Y}}$ denotes the power set of the objective space $\mathcal{Y}$. 
HVI (\autoref{eq:ehvi}) is a popular metric, for instance. 
One advantage of HV is its sensitivity to any type of improvement; whenever an approximation set $A$ dominates another approximation set $B$, then the measure yields a strictly better quality value for the former \citep{zitzler2003performance}. 
On the other hand, HV suffers from sensitivity to transformations of the objectives and scales super-polynomially with $M$, which hinders its practical value.
An alternative approach is to use distance-based indicators \citep{miranda2016necessary, shilton2018multi} that assign scores for the solutions based on a signed distance from
each point to the approximate Pareto front, which is again computationally expensive. 



In the following, the \emph{(weak) Pareto-dominance} relation  is used as a preference relation $\preccurlyeq$ on $\mathcal{Y}$ indicating that a solution ${\bm y'}$ is at least as good as a solution ${\bm y}$ (denoted ${\bm y'} \preccurlyeq {\bm y}$)  iff $f_i({\bm y'}) \leq f_i({\bm y}) \ \forall i \in [M]$.  This relation can be canonically extended to sets of solutions where a set $A \subseteq X$ weakly dominates a set $B \subseteq X$ (denoted $A \preccurlyeq B$)  iff $\forall {\bm y} \in B \ \exists {\bm y'} \in A: {\bm y'} \preccurlyeq {\bm y}$ \citep{zitzler2003performance}. Given the preference relation, we consider the optimization goal of identifying a set of solutions that approximates the set of Pareto-optimal solutions and ideally this set is not strictly dominated by any other approximation set. 

{Since the generalized weak Pareto dominance relation defines only a partial order on $\mathcal{Y}$, there may be incomparable sets in $\mathcal{Y}$. Incomparability is a key challenge in search and performance assessment for multi-objective optimization and becomes more serious as $M$ increases \citep{fonseca2005tutorial}.}
One way to circumvent this problem is to define a total order on $\mathcal{Y}$ which guarantees that any two objective vector sets are mutually comparable. To this end, quality indicators have been introduced that assign, in the simplest case, each approximation set a real number --- that is, a (unary) indicator function $I: \mathcal{Y} \rightarrow \mathbb{R}$ \citep{zitzler2003performance}. One important feature an indicator should have
is \emph{Pareto compliance} \citep{fonseca2005tutorial}, which dictates that it must not contradict the order induced by the
Pareto dominance relation. 

In particular, this means that whenever $A \preccurlyeq B \land  B \nsucceq A$, then the indicator value of A must not be worse than the indicator value of B. A stricter version of compliance would be to require 
that the indicator value of A is strictly better than the indicator value of B (if better means a lower indicator value):
\begin{align}
   A \preccurlyeq B \land  B \nsucceq A \Rightarrow I(A) < I(B).
\end{align}

\subsection{CDF indicator}
We propose the CDF indicator, a Pareto-compliant indicator for measuring the quality of Pareto approximations.
\begin{definition}[Cumulative distribution function]
The CDF of a real-valued random variable $Y$ is the function:\footnote{In this section, we use the standard notation for densities ($f$) and distributions ($F$) {defined on the objective space}. It will be clear from the context whenever $f$ is again used to refer to the objective function.}
\begin{align} \label{def:single_cdf}
F_Y(y) = P(Y \leq y) = \int_{-\infty}^y f_Y(t)dt,
\end{align}
representing the probability that $Y$ takes a value less than or equal to $y$.
\end{definition} 
For more than two variables, the joint CDF is given by
\begin{align} \label{def:mtv_cdf}
& F_{Y_1,\dots, Y_M}({\bm y}) = P(Y_1 \leq y_1, \dots , Y_M \leq y_m) \\
&= \int_{(-\infty, \dots, -\infty)}^{(y_1, \dots, y_M)} f_Y(\mathbf{s})d{\bm s}.
\end{align}

\textbf{Properties of the CDF.}
Every multivariate CDF is monotonically non-decreasing for each $Y_i$, right-continuous in each $Y_i$, and takes values in $[0, 1]$.
The monotonically non-decreasing property means that $F_{{Y}}(a_1, \dots, a_M) \geq  F_{{Y}}(b_1, \dots, b_M)$ whenever $a_1 \geq b_1, \dots, a_K \geq b_M$.  We leverage these properties to define our CDF indicator.
\begin{definition}[CDF Indicator] The CDF indicator ($I_{\rm CDF}$) is defined as the minimum multivariate rank:
\begin{align} \label{def:cdf_indicator}
I_{\rm CDF}(A) &:=
\min_{{\bm y} \in A} F_{{Y}} ({\bm y}) = \max_{{\bm y} \in A} \ [1 - F_{{Y}} ({\bm y})],
\end{align}\label{eq:cdf_score}
where A is an approximation set in $\mathcal{Y}$.
\end{definition}


\begin{theorem}[Pareto compliance] \label{thm:pareto_compliance}
For any pair of approximation sets  $A \in \mathcal{Y} $ and $B \in \mathcal{Y}$,
\begin{align}
A \preccurlyeq B \land  B \nsucceq A \Rightarrow I_{\rm CDF}(A) \leq I_{\rm CDF}(B).
\end{align}
\end{theorem} 
The proof can be found in \autoref{app:app_indicator}.

\begin{remark}
Note that $I_{F_{Y}}$ only depends on the best element in the $F_{Y}$ rank ordering. One consequence of this is that $I_{F_{Y}}$ does not discriminate sets with the same best element. 


\end{remark}

\subsubsection{Estimating the CDF with copulas}

Estimating $F_{Y}$ is challenging in high dimensions.
Naively estimating the joint multivariate density $f_Y$ and then computing the high-dimensional integral to obtain $F_Y$ would be computationally intensive.
To address this, we turn to \emph{copulas} \cite{nelsen2007introduction, Bedford02}, a statistical tool for flexible density estimation in high dimensions. Vine copulas provide consistent factorization of high-dimensional joints into a product of bivariate densities.

\begin{theorem} \label{thm:sklar}
[Sklar's theorem \citep{sklar1959}] The continuous random vector $Y = (Y_1, \dots, Y_M)$ has a joint distribution $F_Y$ and marginal distributions $F_1, \dots, F_M$ iff there exists a unique copula $C$, which is the joint distribution of $U = (U_1, \dots , U_M) =
F_1(Y_1), \dots , F_d(Y_M)$.
\end{theorem}

%

A copula is a multivariate distribution function $C: [0, 1]^M \rightarrow [0, 1]$ that joins (couples) uniform marginal distributions 
$F(y_1, \dots , y_M) = C \left( F_1(y_1), \dots , F_M(y_M) \right)$. To be able to estimate a copula, we need to transform the variables of interest to uniform marginals. We do so by the following operation.


\begin{definition}[Probability integral transform] PIT of a random variable $Y$ with distribution $F_Y$ is the random variable $U = F_Y(Y)$, which is uniformly distributed: $U \sim {\rm Unif}([0, 1])$.
\end{definition}
\autoref{thm:sklar} implies the following corollaries establishing the invariance of the CDF indicator to different scales.

\begin{corollary}
[Scale invariance] A copula based estimator for the CDF indicator is scale-invariant. \label{col1}
\end{corollary}

\begin{corollary}[Invariance under monotonic transformations] Let $Y_1, Y_2$  be continuous random variables with copula $C_{Y_1, Y_2}$. If $\alpha, \beta: \mathbb{R} \rightarrow \mathbb{R}$ are strictly increasing functions, then:
\begin{align} 
C_{\alpha(Y_1), \beta(Y_2)} = C_{Y_1, Y_2},
\end{align}
where $C_{\alpha(Y_1), \beta(Y_2)}$ is the copula function corresponding to variables $\alpha(Y_1)$ and $\beta(Y_2)$. \label{col2}
\end{corollary}

\begin{figure}
    \centering
    \begin{subfigure}
{\includegraphics[trim={0.5cm 0.5cm 0 0.5cm}, clip, width=0.475\textwidth]{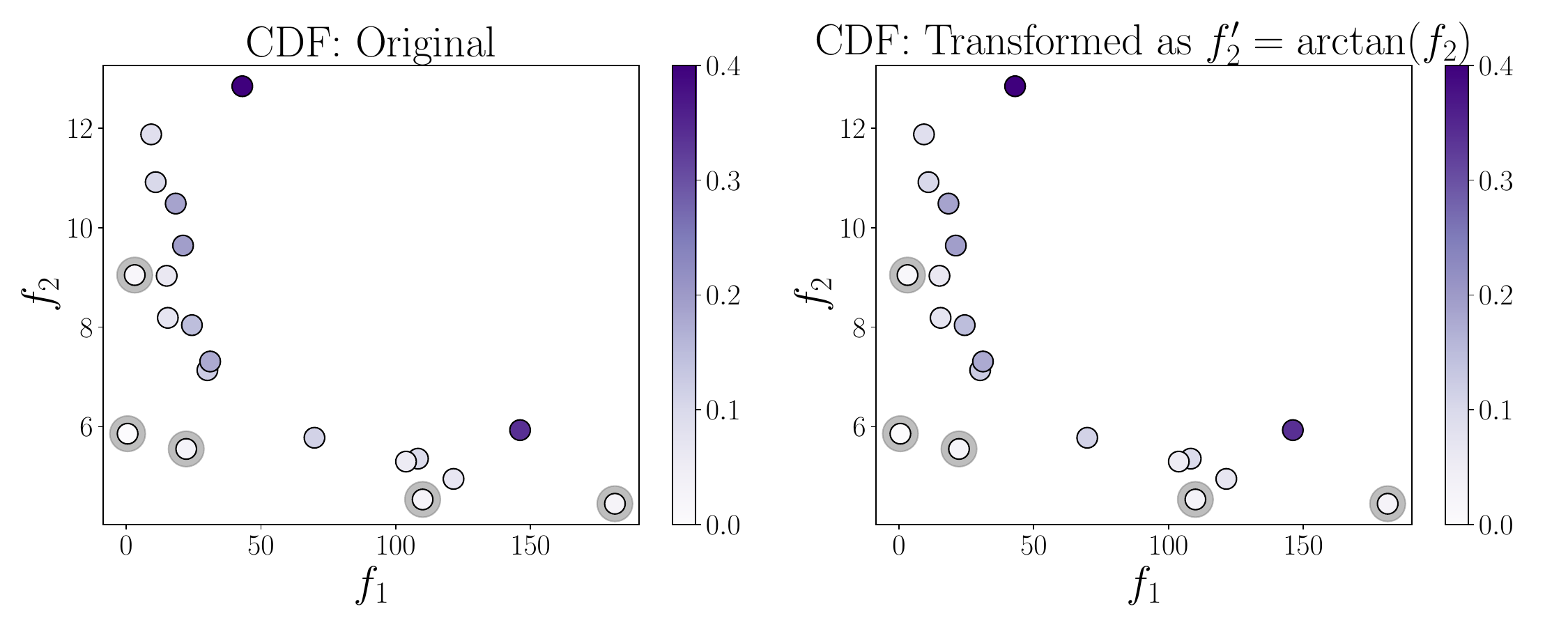}}	
\end{subfigure}
\begin{subfigure}
{\includegraphics[trim={0.5cm 0.5cm 0 0.5cm}, clip, width=0.475\textwidth]{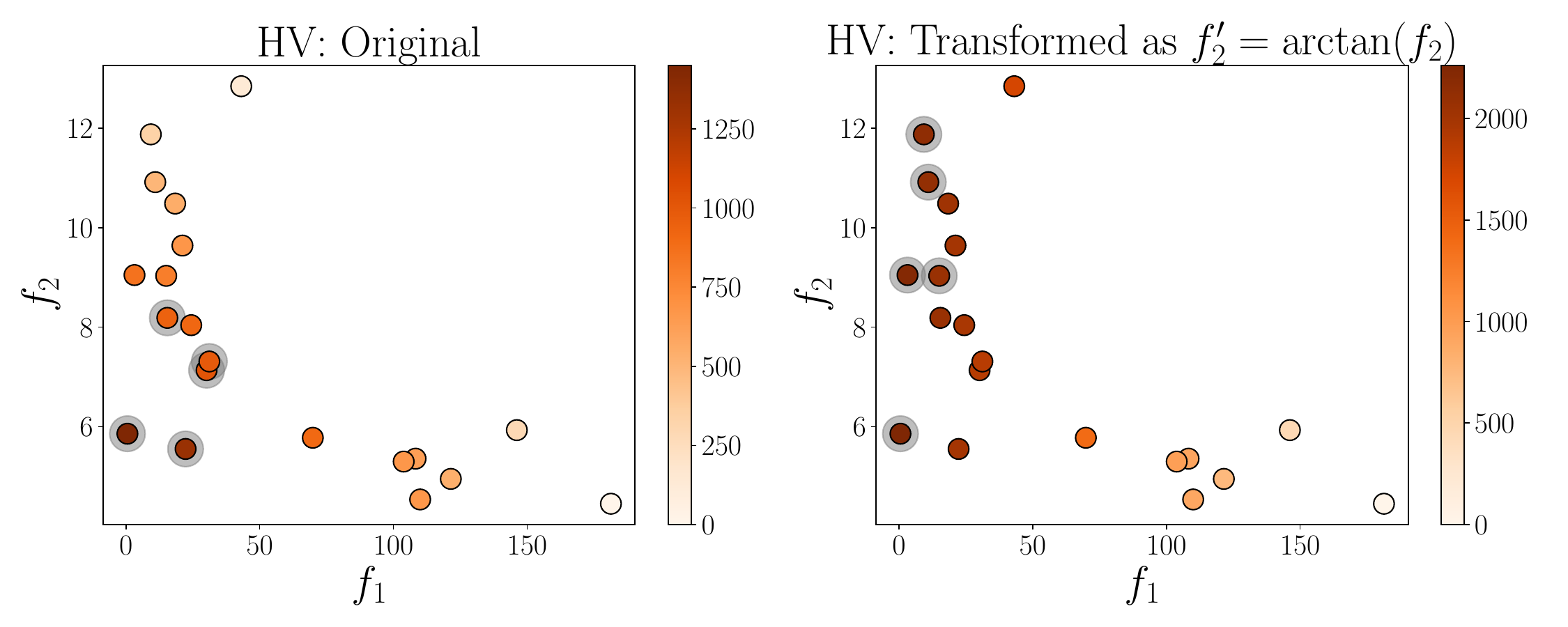}}	
\end{subfigure}
    \caption{Top: The CDF indicator is invariant to arbitrary monotonic transformations of the objectives (here transforming $f_2$ via arctan). Bottom: The HV indicator is highly sensitive to them. The color gradient corresponds to indicator value at each solution ($q=1$). Gray circles are overlaid on the five solutions with the top indicator scores. CDF chooses the same five solutions, but HV prefers ones with high $f_1$ after $f_2$ becomes squashed.}
    \label{fig:metric_compare}
\end{figure}

Corollary \ref{col1} follows from the PIT required for copula estimation. The proof for Corollary \ref{col2}, based on \citet{haugh2016introduction}, can be found in \autoref{app:col2} and, without loss of generality, can be extended to $M{\rm >}2$.
In \autoref{fig:metric_compare} we demonstrate the robustness of the copula-based estimator.

The benefits of using copulas to estimate the CDF are threefold: (i) scalability and flexibility with large $M$, (ii) invariance to relative scales of the different objectives, (iii) invariance to monotonic transformations of the objectives.
\paragraph{From copula density to CDF.} It follows from \autoref{thm:sklar} that a joint density of any bivariate random vector $(Y_1, Y_2)$, can be expressed as $f(y_1, y_2) = f_1(y_1) f_2(y_2) c\left(F_1(y_1), F_2(y_2) \right)$
where $f_1, f_2$ are the marginal densities, $F_1, F_2$ are the marginal distributions, and $c$ is the copula density. In other words, we can factorize the joint density into a product of the marginals and a copula density. The copula density captures the dependence structure between the two variables after all the complexities in the individual margins are removed. 
The factorization speeds up the estimation, which breaks down into two simpler steps: estimating the density of the marginal distributions and estimating the copula density. The parameters of the copula and the margins can be estimated with maximum likelihood given a choice of parametric copula families (lower or upper tail dependence, survival copulas, Gaussian, etc.). In addition, recent progress in nonparametric estimation of copulas has enabled the estimation of more complex distributions \citep{Geenens2017}. Once a copula density is fit, the CDF can be obtained analytically in the parametric case or by Monte-Carlo (MC) integration over the density for the nonparametric case. For further details, please refer to \autoref{app:vc_example}, \citet{aas2009pair}, and \citet{Joe2010}.

\subsubsection{High-dimensional CDF with vine copulas}

The above factorization can be generalized to any number of variables.
The pair copula constructions called \emph{vines} are hierarchical models, constructed from cascades of bivariate copula blocks, that can accommodate more than two variables \cite{nagler2017nonparametric}.
Any $M$-dimensional copula density can be decomposed into a product of ${M(M - 1)}/{2}$ bivariate (conditional) copula densities \citep{Joe97,Bedford02}. The factorization is not unique and can be organized in a graphical model, as a sequence of $M{\rm -}1$ nested trees.
We denote a tree as $T_k = (V_k, E_k)$ with $V_k$ and $E_k$ the sets of nodes and edges of tree $k$ for $k = 1, \dots, M{\rm -}1$. Each edge $e$ is associated with a bivariate copula. We provide a full example of vine copula decomposition in \autoref{app:vc_example}.
In practice, in order to construct a vine, one has to choose two components: 
(1) the structure, or the set of trees $T_k = (V_k, E_k)$ for $k \in [M{\rm -}1]$
and (2) the pair copulas for $ c_{j_e, k_e | D_e}$ where $e \in E_k$ and $k \in [M{\rm -}1]$.
There are efficient algorithms for both steps and we use the implementation by \citet{Nagler2018}. 

\subsection{CDF-based acquisition function: BOtied}
Suppose we fit a CDF on ${\bm y}^{(1)}, {\bm y}^{(2)}, \dotsc, {\bm y}^{(N_t)}$, the $N_t$ measurements acquired thus far. Denote the resulting CDF as ${\hat F}(\cdot; \mathcal{D}_t)$, where we have made explicit the dependence on the dataset up to time $t$. The utility function of our {BOtied} acquisition function is as follows:
\begin{align}\label{eq:utility_botied}
    u({\bf x}, {\hat f}, \mathcal{D}_t) = 1 - {\hat F}( {\hat f}({\bf x}); \mathcal{D}_t ).
\end{align}
As with the CDF indicator, our CDF-based acquisition function has an efficient implementation based on vine copulas. 
For a more precise description of how a CDF-based acquisition function fits within a single round of MOBO, we include Algorithm \ref{alg:algo} in \autoref{app:algorithm}. 

 \begin{figure}[t!]
     \centering
\includegraphics[width=0.47\textwidth]{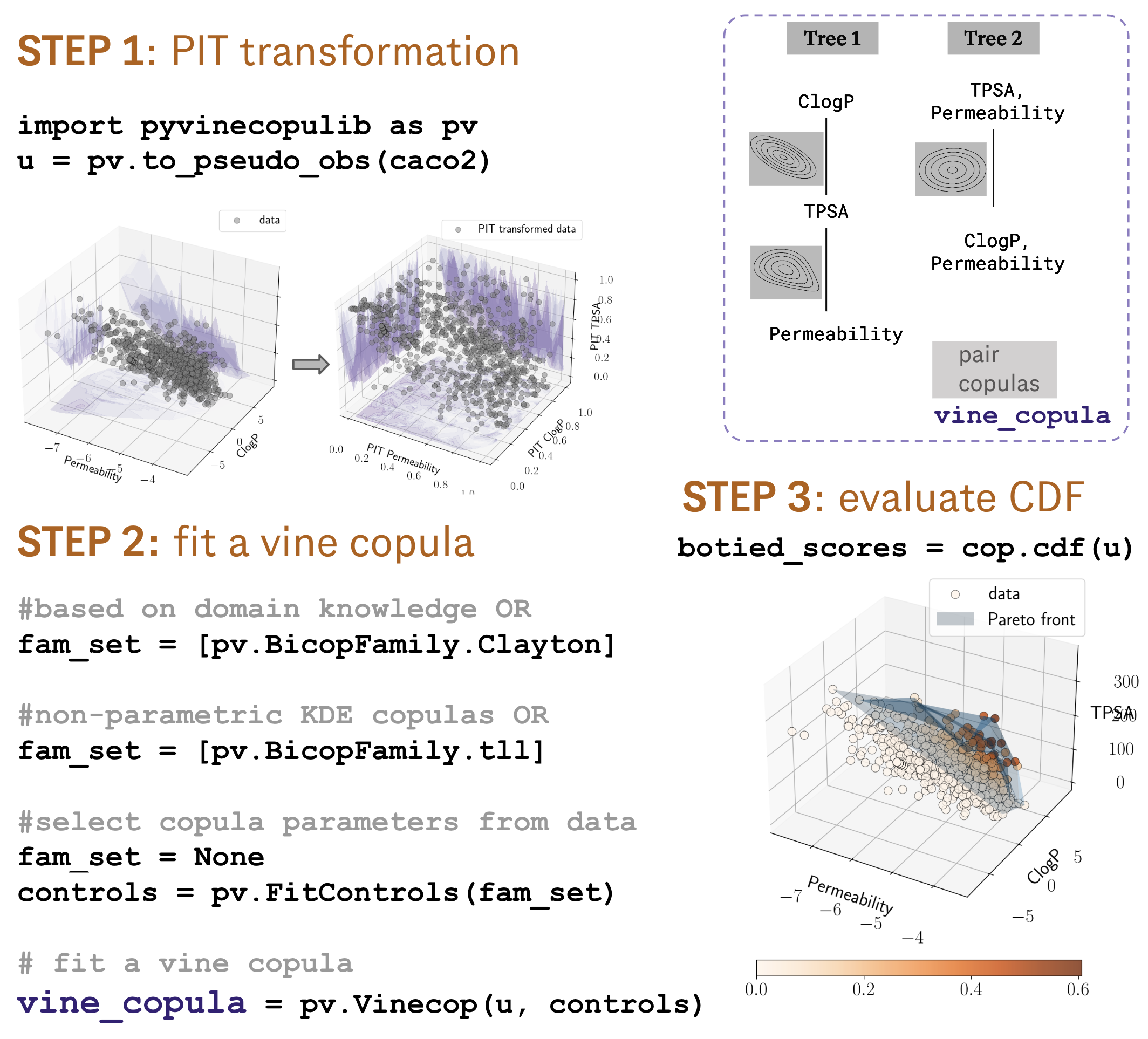}
\vspace{-0.3cm}
     \caption{A recipe for estimating the CDF with copulas, in three simple steps and fewer than 5 lines of Python code. Plots are based on the Caco2+ dataset.}
     \label{fig:vc_impl}
 \end{figure}

\section{Empirical results}
\label{sec:experiments}

\begin{table*}[ht]
\centering
\caption{HV indicators (in the original units) and $I_{\rm CDF}$ across datasets. Higher is better for HV and lower is better for $I_{\rm CDF}$. The best per column is marked in bold. We report the mean and standard error of each metric across 20 random seeds.} \label{tab:toy}
\begin{tabular}{lllllll}
          \hline
          & \multicolumn{2}{l}{Penicillin ($d=7, M=3, q=1$)}     & \multicolumn{2}{l}{DTLZ2 ($d=6, M=4, q=1$)}       & \multicolumn{2}{l}{DTLZ2 ($d=7, M=6, q=1$)}       \\ 
          \hline
          & $I_{\rm CDF}$ $\downarrow$ & HV $\uparrow$           & $I_{\rm CDF}$ $\downarrow$ & HV $\uparrow$        & $I_{\rm CDF}$ $\downarrow$ & HV $\uparrow$        \\
          \hline
BOtied v1 & 0.26 (0.01)                & 325741 (29515)          & 0.25 (0.01)                & 2.26 (0.09)          & 0.071 (0.005)              & 0.36 (0.03)          \\
BOtied v2 & \textbf{0.20 (0.02)}       & \textbf{342762 (13599)} & 0.11 (0.02)                & \textbf{2.32 (0.06)} & 0.064 (0.004)              & \textbf{0.42 (0.02)} \\
NParEGO   & 0.28 (0.01)                & 303707 (15118)          & \textbf{0.10 (0.02)}       & 2.20 (0.11)          & 0.065 (0.005)              & 0.38 (0.02)          \\
NEHVI     & 0.28 (0.01)                & 314294 (14498)          & 0.24 (0.01)                & 1.80 (0.06)          & 0.074 (0.007)              & 0.27 (0.01)          \\
PES       & 0.27 (0.01)                & 297107 (17383)          & 0.24 (0.01)                & 1.85 (0.11)          & 0.069 (0.004)              & 0.23 (0.06)          \\
MES       & 0.24 (0.02)                & 305874 (14694)          & \textbf{0.10 (0.02)}       & 2.12 (0.08)          & \textbf{0.059 (0.004)}     & 0.27 (0.06)          \\
JES       & 0.28 (0.01)                & 316302 (21193)          & 0.24 (0.01)                & 1.97 (0.07)          & 0.069 (0.006)              & 0.25 (0.05)          \\
Random    & 0.24 (0.02)                & 307896 (22889)          & 0.11 (0.02)                & 0.91 (0.08)          & 0.076 (0.005)              & 0.10 (0.01)         
\end{tabular}
\end{table*}

\textbf{Experimental setup.}
To empirically evaluate the sample efficiency of {BOtied}, we execute simulated BO rounds on a variety of problems. See \autoref{app:experimental_detail} for more details about our setup. For all the experiments, the surrogate model was an independent GP with a Matern 5/2 ARD kernel. The GP hyperparameters were inferred via maximum a posteriori (MAP) estimation. The code that reproduces all of our experiments and plots is available at \url{https://github.com/jiwoncpark/botied} \faGithub. 

\begin{figure*}[ht!]
\begin{center}
\includegraphics[width=0.95\textwidth]{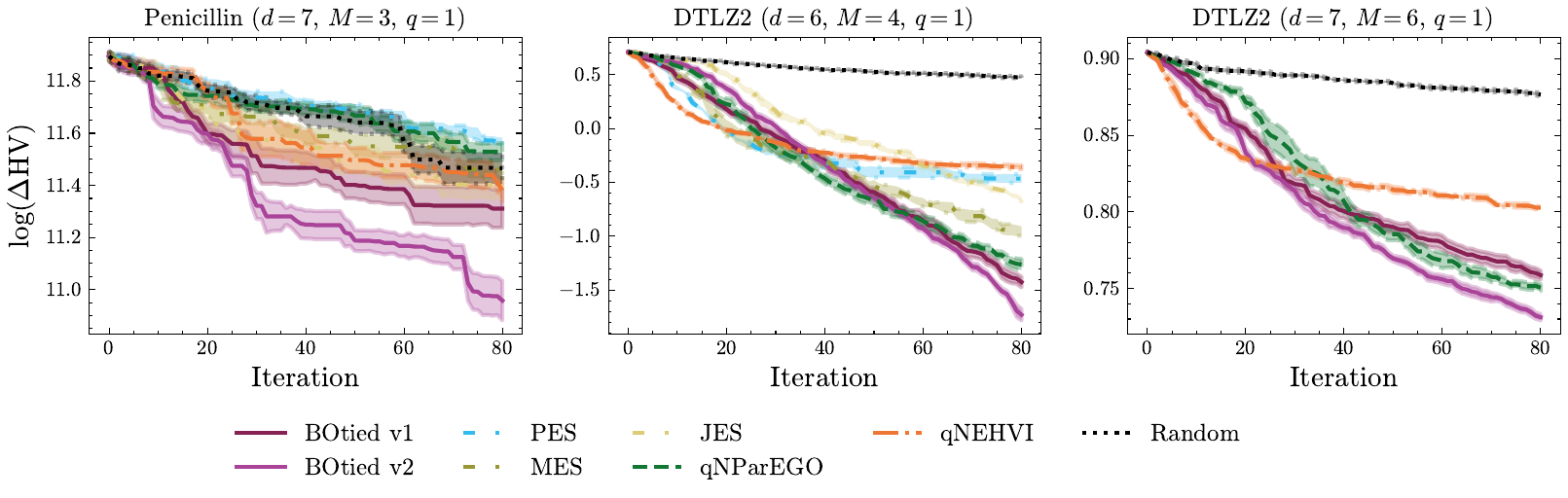}
\end{center}
\vspace{-0.5cm}
\caption{HV vs. iterations for three synthetic test functions. We show the mean and two standard errors over 20 random seeds. \label{fig:metric_vs_iters}}
\end{figure*}

\textbf{Metrics.} We use the HV indicator presented in 
\autoref{sec:method}, a standard evaluation metric for MOBO, as well as our CDF indicator $I_{\rm CDF}$ on the noiseless function values. We rely on efficient algorithms for HV computation based on hyper-cell decomposition as described in \citep{fonseca2006improved, ishibuchi2011many} and implemented in \verb|BoTorch| \citep{balandat2020botorch}.

\textbf{BOtied implementation} We implement two versions of {BOtied} that differ in the incorporation of predictive uncertainties in the CDF estimation. In one version (v1), we fit the CDF on all of the MC predictive posterior samples across all the candidates. This can sometimes result in poor CDF fit, particularly when uncertainties are large. The other version (v2) alleviates this issue by fitting the CDF on the posterior means of the candidates. The algorithms for both versions can be found in Algorithm \ref{alg:algo}, \autoref{app:algorithm}. 
We optimize the {BOtied} acquisition values using the gradient-free CMA-ES algorithm \citep{hansen2006cma}. The CDF estimation is detailed in \autoref{app:other_cdf_scores} and {BOtied} optimization in \autoref{app:experimental_detail}. 

\textbf{Baselines} We compare {BOtied} with the noisy versions of popular acquisition functions. The baseline acquisition strategies are NEHVI \citep{daulton2020differentiable} described in \autoref{eq:nehvi}; noisy NParEGO \citep[NParEGO;][]{knowles2006parego} which uses noisy EI on top of random augmented Chebyshev scalarization; predictive entropy search \citep[PES;][]{hernandez2016predictive}, maximum entropy search \citep[MES;][]{belakaria2019max}, and joint entropy search \citep[JES;][]{hvarfner2022joint} --- the differences being the estimation of entropy in the inputs, objectives, or both, respectively; and {random} (Sobol) selection.

\textbf{Synthetic datasets.} We include synthetic test functions for direct evaluation of $f$. We focus on ones that support $M\geq 3$: DTLZ2 \citep[$d{\rm =}6, M{\rm =}4$ and $d{\rm =}7, M{\rm =}6$;][]{deb2005searching} and Penicillin \citep[$d{\rm =}7, M{\rm =}3$;][]{liang2021scalable}, which simulates the penicillin yield, time to production, and undesired byproduct for various parameters of the production process. See \autoref{app:experimental_detail} for more detail. 

\textbf{Real-world datasets.} To emulate a multi-objective drug design setting, we postprocess the real-world dataset Caco2 \cite{wang2016adme} from the Therapeutics Data Commons database \citep{Huang2021tdc,Huang2022artificial} to create Caco2+. The original Caco2 dataset consists of 906 drug molecules annotated with experimentally measured cell permeability, or the rates of passing through a human colon epithelial cancer cell line. Permeability is a key property in the absorption, distribution, metabolism, and excretion (ADME) profile of drugs. We augment the dataset with additional properties using RDKit \citep{landrum_rdkit}, including ClogP related to fat solubility
and 
topological polar surface area (TPSA).
Subsets of these properties (e.g., permeability and TPSA) are inversely correlated and thus compete with one another during optimization.
In late-stage lead-molecule optimization, the trade-offs become more dramatic and as more properties are added \citep{sun202290}. Demonstrating effective sampling of Pareto-optimal solutions in this setting is thus of great value. We represent each molecule as a concatenation of fingerprint and fragment feature vectors \citep{thawani2020photoswitch}. 

We also include experiments over three datasets from the DDMOP benchmark \citep{ddmop}. Differently from the synthetic test functions which have analytical solutions, each DDMOP dataset represents a complex objective function approximated by expensive numerical simulations. These datasets address cab car optimization, power system chip placement and neural network. Details and table results on each dataset can be found in \autoref{app:ddmop}. See \autoref{app:experimental_detail} for more detail about these datasets.

 \textbf{Vine Copulas for MOBO in practice}
 In \autoref{fig:vc_impl}, we present a simple recipe for estimating CDFs with vine copulas, in three simple steps and fewer than five lines of code. We use the Caco2+ dataset ($M{\rm=}3$) as an example. First, the PIT transformation yields the uniform margins. We then choose a copula shape from parametric or non-parametric families, or we leave this undetermined and run model selection based on the Bayesian information criterion (BIC). In the case of Caco2+, we can use the domain knowledge that permeability and TPSA are negatively correlated and specify a Clayton copula. Once a vine copula has been fit, it is fully described by the trees (structure) and bivariate (pair) copula densities associated with each edge. This is all we need to evaluate the CDF on the data points in Step 3. Note that the darker shaded points corresponding to higher CDF scores indeed approach the Pareto front. 

\subsection{Results and discussion}

\begin{figure}
\begin{center}
\includegraphics[width=0.46\textwidth]{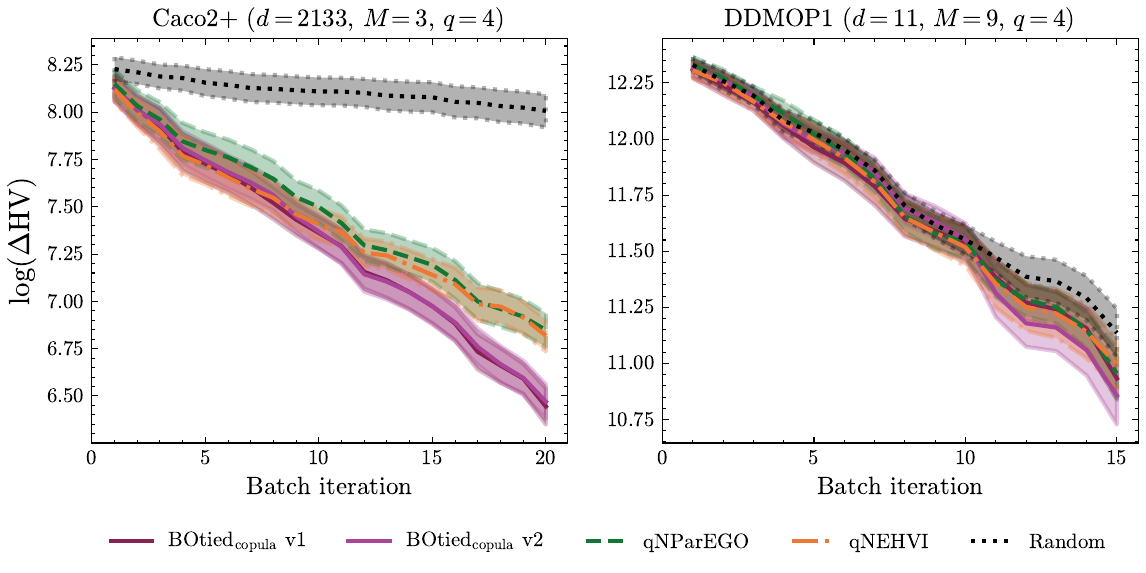}
\end{center}
\vspace{-0.5cm}
\caption{HV vs. iterations for real-world datasets. We show the mean and two standard errors over 20 random seeds. \label{fig:metric_vs_iters_real}}
\end{figure}

We compare the performance of {BOtied} with baseline acquisition strategies in terms of both the HV and the CDF indicators, on synthetic test functions (\autoref{fig:metric_vs_iters}) as well as on real-world datasets (\autoref{fig:metric_vs_iters_real}). The metrics for these experiments and additional experiments using various $q$ batch sizes are tabulated in \autoref{tab:toy}. Although there is no single best method across all the datasets, the best numbers are consistently achieved by either {BOtied} v1 or v2 with NParEGO being a close competitor. The NEHVI performance visibly degrades as $M$ increases.

\begin{figure}
  \begin{center}
    \includegraphics[trim={0.25cm, 0.6cm, 0cm, 1cm}, clip, width=0.41\textwidth]{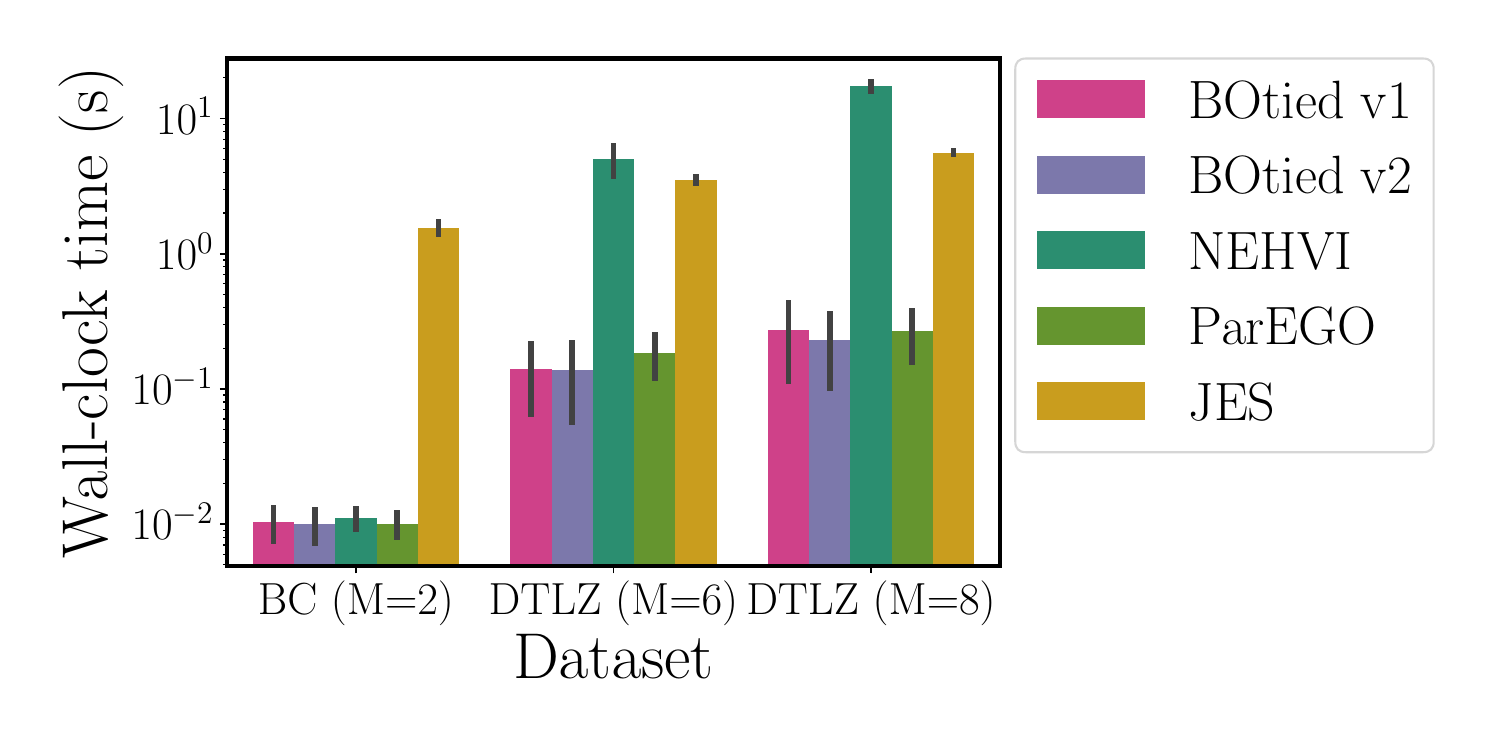}
  \end{center}
  \vspace{-0.5cm}
  \caption{Wall-clock time per single call of acquisition function. Error bars are standard deviations across five repeated calls. \label{fig:time}}
\end{figure}

\autoref{fig:time} shows that the wall-clock time for NEHVI and JES become very slow for $M{\rm \geq}3$.
At the same time, {BOtied} is significantly faster than NEHVI/JES and is as fast as NParEGO, which is based on scalarizing the $M$ objectives.

There are two main benefits to using $I_{\rm CDF}$ rather than HV for evaluation. First, the CDF is bounded between 0 and 1, with scores close to 0 corresponding to the solutions closest to our approximate Pareto front. Unlike HV values, for which the scales do not carry information about the internal ordering, the $I_{\rm CDF}$ values have an interpretable scale. Second, assuming the GP and copula have been properly fit, we can use the magnitude of $I_{\rm CDF}$ to determine the orthogonality, or degree of competition, of the objectives in a given task. In particular, when a candidate strongly dominates a set of points, its $I_{\rm CDF}$ tends below 0.1, while for points that weakly dominate with respect to a small subset of the objectives, the $I_{\rm CDF}$ value is higher. 

We stress-test {BOtied} in a series of ablation studies in \autoref{app:ablation}. In particular, we vary the number of MC posterior predictive samples and find that {BOtied} v1 is robust to the number of posterior samples, i.e., the multivariate ranks associated with the best-fit copula model do not change significantly with varying numbers of samples. When the posterior shape is complex such that many MC samples are required to fully characterize the posterior, {BOtied} v2 (in which the copula is fit on the mean of the posterior samples) is more appropriate than v1. 
\vspace{-0.5cm}
{\paragraph{Limitations} When fitting the CDF model, there's a trade-off between flexibility and complexity. Increasing $M$ requires us to adopt more flexible models, which increases the number of modeling choices. In our experiments, we perform model selection based on the Akaike information criterion (AIC) to choose among nonparametric and parametric copula families \citep{akaike1998information}.
Moreover, the current implementation of BOtied is not differentiable, which necessitates the use of gradient-free algorithms such as CMA-ES for optimizing acquisition values where a gradient-based one may be more efficient. 
} 

\section{Conclusion}
\label{sec:conclusion}
We introduce a new perspective on MOBO based on the multivariate CDF. Our proposed MOBO acquisition function, {BOtied}, is computed by fitting a multivariate CDF on the surrogate predictions and extracting the ranks associated with the CDF scores. It is computationally attractive, as the CDF can be efficiently fit with vine copulas even when $M$ is large.  
Moreover, it enables model-based estimation of the Pareto front. When domain knowledge about the distribution of the objective values is available, it can be injected into the specification of the CDF model family.
We also propose a new Pareto-compliant indicator for measuring the quality of approximate Pareto fronts, the CDF indicator. The CDF indicator, equipped with desirable properties such as invariance to monotonic transformations of the objectives, promises to complement the popular HV indicator.  

Our method is general and lends itself to a number of immediate extensions. 
First, whereas we have implemented gradient-free optimization of {BOtied} in this work, we can take advantage of gradient-based optimization for improved efficiency. In conjunction with a differentiable sorting algorithm \citep[e.g.,][]{cuturi2019differentiable,blondel2020fast}, the computation of our acquisition function can be made differentiable for many parametric copula families.
Second, we can consider {constrained or discrete extensions} for broader applicability.
Finally, as many applications carry noise in the input as well as the function of interest, accounting for input noise through the established connection between copulas and multivariate value-at-risk (MVaR) estimation will be of great practical interest.

\section*{Impact Statement}


This paper presents work whose goal is to advance the field of 
Machine Learning. There are many potential societal consequences 
of our work, none which we feel must be specifically highlighted here.



\bibliography{main}
\bibliographystyle{icml2024}

\newpage
\appendix
\onecolumn

\section{Related work in multi-objective optimization}
\label{app:related_work}
\begin{table}[H]
\centering
\small
\caption{Comparison of BOtied with related work}
\adjustbox{max width=0.9\textwidth}{
\begin{tabular}{@{}l|r|ccccc@{}}
\toprule
\multirow{2}{*}{\textbf{\begin{tabular}[c]{@{}l@{}}Type of \\ groundwork\end{tabular}}} &
  \multirow{2}{*}{\textbf{\begin{tabular}[c]{@{}r@{}}Scoring \\ method\end{tabular}}} &
  \multicolumn{1}{l}{\multirow{2}{*}{\textbf{\begin{tabular}[c]{@{}l@{}}MO \\ criteria\end{tabular}}}} &
  \multicolumn{1}{l}{\multirow{2}{*}{\textbf{\begin{tabular}[c]{@{}l@{}}Scalability \\ with M\end{tabular}}}} &
  \multicolumn{1}{l}{\multirow{2}{*}{\textbf{\begin{tabular}[c]{@{}l@{}}Scale \\ invariance\end{tabular}}}} &
  \multicolumn{1}{l}{\multirow{2}{*}{\textbf{\begin{tabular}[c]{@{}l@{}}Bayesian \\ optimization\end{tabular}}}} &
  \multicolumn{1}{l}{\multirow{2}{*}{\textbf{\begin{tabular}[c]{@{}l@{}}Non GP\\ surrogates\end{tabular}}}} \\
 &
   &
  \multicolumn{1}{l}{} &
  \multicolumn{1}{l}{} &
  \multicolumn{1}{l}{} &
  \multicolumn{1}{l}{} &
  \multicolumn{1}{l}{} \\ \midrule \midrule
\textbf{\begin{tabular}[c]{@{}l@{}}Multivariate ranks/CDF, \\ Copula, Copula space \end{tabular}} &
  \textbf{\begin{tabular}[c]{@{}r@{}}BOtied\\ (this work)\end{tabular}} &
  \cmark &
  \cmark &
  \cmark &
  \cmark &
  \cmark \\ \cmidrule(r){1-2} 
\textbf{\begin{tabular}[c]{@{}l@{}}Copula space, \\ Game theory\end{tabular}} &
  \textbf{\begin{tabular}[c]{@{}r@{}}Kalai-Smorodinsky MO \citep{binois2020kalai}\end{tabular}} &
  \cmark & \cmark
   &
  \cmark &
  \cmark &
  \xmark \\ \cmidrule(r){1-2}
\multirow{2}{*}{\textbf{\begin{tabular}[c]{@{}l@{}}Multivariate \\ ranks\end{tabular}}} &
  \textbf{\begin{tabular}[c]{@{}r@{}}Aggregate Rank \citep{kukkonen2007ranking}\end{tabular}} &
  \cmark &
  \cmark &
  \cmark &
  \xmark &
  \xmark 
  \\
 &
  \textbf{\begin{tabular}[c]{@{}r@{}}Ordinal BO \citep{picheny2019ordinal}\end{tabular}} &
  \cmark &
  \xmark &
  \xmark &
  \cmark &
  \cmark \\ \cmidrule(r){1-2}
\multirow{3}{*}{\textbf{\begin{tabular}[c]{@{}l@{}} \\ Information \\ Theoretic\end{tabular}}} &
  \textbf{\begin{tabular}[c]{@{}r@{}}Joint Entropy Search\\ \citep{tu2022joint, hvarfner2022joint}\end{tabular}} &
  \cmark &
  \xmark &
  \cmark &
  \cmark &
  \xmark \\
 &
  \textbf{\begin{tabular}[c]{@{}r@{}}Predictive Entropy Search \citep{hernandez2016predictive}\end{tabular}} &
  \cmark &
  \xmark &
  \xmark &
  \cmark &
  \xmark \\
 &
  \textbf{\begin{tabular}[c]{@{}r@{}}Max-Value Entropy Search \citep{belakaria2019max}\end{tabular}} &
  \cmark &
  \xmark &
  \xmark &
  \cmark &
  \xmark \\ \cmidrule(r){1-2}
\textbf{Hypervolume} &
  \textbf{\begin{tabular}[c]{@{}r@{}}EHVI variants \citep{daulton2021parallel, daulton2022bayesian}\end{tabular}} &
  \cmark &
  \xmark &
  \xmark &
  \cmark &
  \cmark \\ \cmidrule(r){1-2}
\textbf{\begin{tabular}[c]{@{}l@{}}Random\\ scalarization\end{tabular}} &
  \textbf{\begin{tabular}[c]{@{}r@{}}ParEGO \citep{knowles2006parego}\end{tabular}} &
  \cmark &
  \cmark &
  \xmark &
  \cmark &
  \cmark \\ \cmidrule(r){1-2}
\textbf{\begin{tabular}[c]{@{}l@{}}Boundary \\ distance\end{tabular}} &
  \textbf{\begin{tabular}[c]{@{}r@{}}SVM-variants \citep{miranda2016necessary, shilton2018multi}  \end{tabular}} &
  \cmark &
  \cmark &
  \xmark &
  \cmark &
  \xmark \\
  \cmidrule(r){1-2}
  \textbf{\begin{tabular}[c]{@{}l@{}}Maxmin, Pareto Indicator\end{tabular}} &
  \textbf{\begin{tabular}[c]{@{}r@{}}Pareto improvement , EmaX \citep{bautista2009sequential}\\	Maximin improvement \citep{svenson2011computer}  \end{tabular}} &
  \cmark &
  \cmark &
  \xmark &
  \cmark &
  \cmark \\ \cmidrule(r){1-2}
    \textbf{\begin{tabular}[c]{@{}l@{}}Completeness\end{tabular}}  &
   \textbf{\begin{tabular}[c]{@{}r@{}}Averaged completeness indicator \citep{svenson2011computer}\end{tabular}} &
  \cmark &
  \xmark &
  \cmark &
  \cmark &
  \xmark  \\
 &
  \textbf{\begin{tabular}[c]{@{}r@{}}Estimated completeness indicator improvement \citep{svenson2011computer}\end{tabular}} &
  \cmark &
  \cmark &
  \cmark &
  \cmark &
  \cmark \\ \bottomrule \bottomrule
  
\end{tabular}}
\label{tab:overview}
\end{table}

\section{Algorithm}
\label{app:algorithm}

\begin{minipage}{0.9\textwidth}
\centering
\begin{algorithm}[H]
\centering
\caption{MOBO with {BOtied}: a CDF-based acquisition function}
\begin{algorithmic}[1]
\label{alg:algo}
\STATE \textbf{Input:} Surrogate model ${\hat f}$, initial data $\mathcal{D}_0 = \{({\bm x}_n, {\bm y}_n)\}_{n=1}^{N_0}$, $\mathcal{X} \subset \mathbb{R}^d, \mathcal{Y} \subset \mathbb{R}^M$, number of MOBO iterations $T$, size of the candidate pool used in each inner-loop optimization of the acquisition function $N$, number of posterior predictive sample $L$\\
\STATE \textbf{Output:} Optimal selected subset $\mathcal{D}_T$.


\FOR{$\lbrace t = 1, \dots, T \rbrace$}
\WHILE{converged}
\STATE {Sample the candidate pool $X \coloneqq [{\bm x}_1, \cdots, {\bm x}_N] \subset \mathcal{X}$}

\STATE {Obtain the predictive distribution $p(f|\mathcal{D}_{t-1}, X)$} \\
\STATE {Draw $L$ predictive samples ${\hat f}^{(j)} \sim p(f|\mathcal{D}_{t-1}, X)$}, for $j\in [L]$

\STATE{Version 1: Fit a CDF $\hat F$ on the pooled samples, $\{ {\hat f}^{(j)} \}_{j\in [L]}$. \\
Version 2: Fit a CDF $\hat F$ on the mean-aggregated samples, $\frac{1}{L}\sum_{j=1}^L {\hat f}^{(j)}$ (or posterior mean parameters if they are directly available from the parameterization of the $\hat f$ posterior).
}
\vspace{1mm}
\FOR{$\lbrace i = 1, \dots, N \rbrace$}
    \STATE {Version 1: Evaluate the fit CDF $\hat F$ on the samples and take the mean across the samples $\mathcal{S}({\bm x}_i) = \frac{1}{L}\sum_{j=1}^L {\hat C} \left( {\hat f}_i^{(j)} \right)$ \\
    Version 2: Evaluate the fit CDF $\hat F$ on the posterior means $\mathcal{S}({\bm x}_i) = {\hat C}\left( \frac{1}{L}\sum_{j=1}^L {\hat f}_i^{(j)} \right)$
    }
\ENDFOR
\ENDWHILE
\STATE{
$i^\star \leftarrow \argmax_{i \in [N]}\mathcal{S}({\bm x}_i)$
} \\
\STATE{
$\mathcal{D}_t \leftarrow \mathcal{D}_{t-1} \cup \lbrace ({\bm x}_{i^\star}, {\bm y}_{i^\star}) \rbrace$
}
\ENDFOR \\
{\bf return} {$\mathcal{D}_T$}
\end{algorithmic}
\end{algorithm}
\end{minipage}

\section{Properties of the CDF indicator}
\label{app:app_indicator}
\subsection{Theorem 1: Pareto compliance of the CDF indicator} 

We state \autoref{thm:pareto_compliance} again and provide the proof here. 

\textbf{\autoref{thm:pareto_compliance}}: For any arbitrary approximation sets $A \in \mathcal{X}$ and $B \in \mathcal{X}$ where $\mathcal{X} \subset \mathbb{R}^d$, the following holds:
\begin{align*}
A \preccurlyeq B \land  B \nsucceq A \Rightarrow I_{F}(A) \leq I_{F}(B).
\end{align*}

\begin{proof}
If we have $A \preccurlyeq B \land  B \nsucceq A$, then the following two conditions hold: $\forall {\bm x}' \in B \ \exists {\bm x} \in A: {\bm x} \preccurlyeq {\bm x}'$ and $\exists \mathbf{x} \in A \ s.t. \ \nexists {\bm x}' \in B: {\bm x}' \preccurlyeq \mathbf{x}$. Recall that the weak Pareto dominance $\mathbf{x} \preccurlyeq \mathbf{x}'$ implies that $\forall i \in [M]: f_i({\bm x}) \leq f_i({\bm x}')$. From the definition and fundamental property of the CDF being a monotonic non-decreasing function, it follows that $\forall i \in [M]: f_i({\bm x}) \leq f_i({\bm x}') \Rightarrow F_Y({\bm x}) \leq F_ Y({\bm x}')$. 

Define the set of non-dominated solutions in $B$, $\mathcal{P}_B \coloneqq \{{\bm x} \in B, \forall {\bm x}' \in B : {\bm x} \preceq {\bm x}'\}$. Note that $I_{F}(B) = I_F(\mathcal{P}_B) = I_F(\{{\bm z}\})$ for any ${\bm z} \in \mathcal{P}_B$. Now let ${\bm x}_B \in \mathcal{P}_B$. There is ${\bm x}_A \in A$ such that ${\bm x}_A \preceq {\bm x}_B$, and we have that $F_Y({\bm x}_A) \leq F_ Y({\bm x}_B)$. By definition, $I_F(A) \leq I_F(\{{\bm x}_A\})$ so we have $I_{F}(A) \leq I_{F}(\{{\bm x}_A\}) \leq I_{F}(\{{\bm x}_B\}) = I_F (B)$ as desired.
\end{proof}

\subsection{Corollary 2: Invariance under monotonic transformations} \label{app:col2}
This proof closely follows the one in \citep{haugh2016introduction}.

\textbf{Corollary 2:} 
Let $Y_1, Y_2$  be continuous random variables with copula $C_{Y_1, Y_2}$. If $\alpha, \beta: \mathbb{R} \rightarrow \mathbb{R}$ are strictly increasing functions, then:
\begin{align}
C_{\alpha(Y_1), \beta(Y_2)} = C_{Y_1, Y_2}
\end{align}
where $C_{\alpha(Y_1), \beta(Y_2)}$  is the copula function corresponding to variables $\alpha(Y_1)$ and $\beta(Y_2).$

\begin{proof}
    We first note that for the distribution function of $\alpha(Y_1)$ it holds that
    \begin{align}
        F_{\alpha(Y_1)} = P(\alpha(Y_1) \leq y_1) = P(Y_1 \leq \alpha^{-1}(y_1)) = F_{Y_1}(\alpha^{-1}(y_1))
    \end{align}\label{eq:alpha}
    and analogously,
    \begin{align}
        F_\beta(Y_1) (y_1) = F_{Y_1}(\beta^{-1}(y_1))
    \end{align}
    \label{eq:beta}

    From Sklar's theorem, we have that for all $y_1, y_2 \in \mathbb{R}$

\begin{align*}
    C_{\alpha(Y_1)\beta(Y_2)}(F_{\alpha(Y_1)}(y_1), F_{\beta(Y_2)}(y_2)) 
    &=  F_{\alpha(Y_1)\beta(Y_2)}(y_1, y_2) \\
    &= P(\alpha(Y_1) \leq y_1, \beta(Y_2) \leq y_2)  \\
    &= P(Y_1 \leq \alpha^{-1}(y_1), Y_2 \leq \beta^{-1}(y_2))  \\
    &= F_{Y_1, Y_2}(\alpha^{-1}(y_1), \beta^{-1}(y_2)) ) \\
    &= C_{Y_1, Y_2} (F_{Y_1}(\alpha^{-1}(y_1)), F_{Y_2}(\beta^{-1}(y_2))) \\ &= C_{Y_1, Y_2} (F_{\alpha(Y_1)}(y_1),F_{\beta(Y_2)}(y_2))
\end{align*}
Equalities one and five follow from Sklar's theorem. In the third equality we make use of fact that $\alpha$ and $\beta$ are increasing functions. The last equality follows from \autoref{eq:alpha} and \autoref{eq:beta}.
\end{proof}


\section{(Vine) copula overview and example} \label{app:vc_example}
 
According to  Sklar’s theorem \cite{sklar1959}, the joint density of any bivariate random vector $(X_1, X_2)$, can be expressed as
\begin{align}\label{eq:cop}
    f(x_1, x_2) = f_1(x_1)f_2(x_2) c\left(F_1(x_1), F_2(x_2) \right)
\end{align}
where $f_i$\footnote{In this section, we use the standard notations for densities ($f$) and distributions ($F$) as commonly done in the copula literature.} are the marginal densities, $F_i$ the marginal distributions, and $c$ the copula density. 

That is, any bivariate density is uniquely described by the product of its marginal densities and a \emph{copula density}, which is interpreted as the \emph{dependence structure}. For self-containment of the manuscript, we borrow an example from \cite{tagasovska2023retrospective}. \autoref{fig:copula}.7 illustrates all of the components representing the joint density. 

As a benefit of such factorization, by taking the logarithm on both sides, one can estimate the joint density in two steps, first for the marginal distributions, and then for the copula. 
Hence, copulas provide a means to flexibly specify the marginal and joint distribution of variables. For further details, please refer to \cite{aas2009pair, Joe2010}.

There exist many parametric representations through different copula families, however, to leverage even more flexibility, in this paper, we focus on the kernel-based nonparametric copulas of \cite{Geenens2017}.

\autoref{eq:cop} can be generalized and holds for any number of variables.
To be able to fit densities of more than two variables, we make use of the pair copula constructions, namely \emph{vines}; hierarchical models, constructed from cascades of bivariate copula blocks \cite{nagler2017nonparametric}.
\begin{wrapfigure}{r}{0.43\textwidth}
  \begin{center}
    \includegraphics[width=0.37\textwidth]{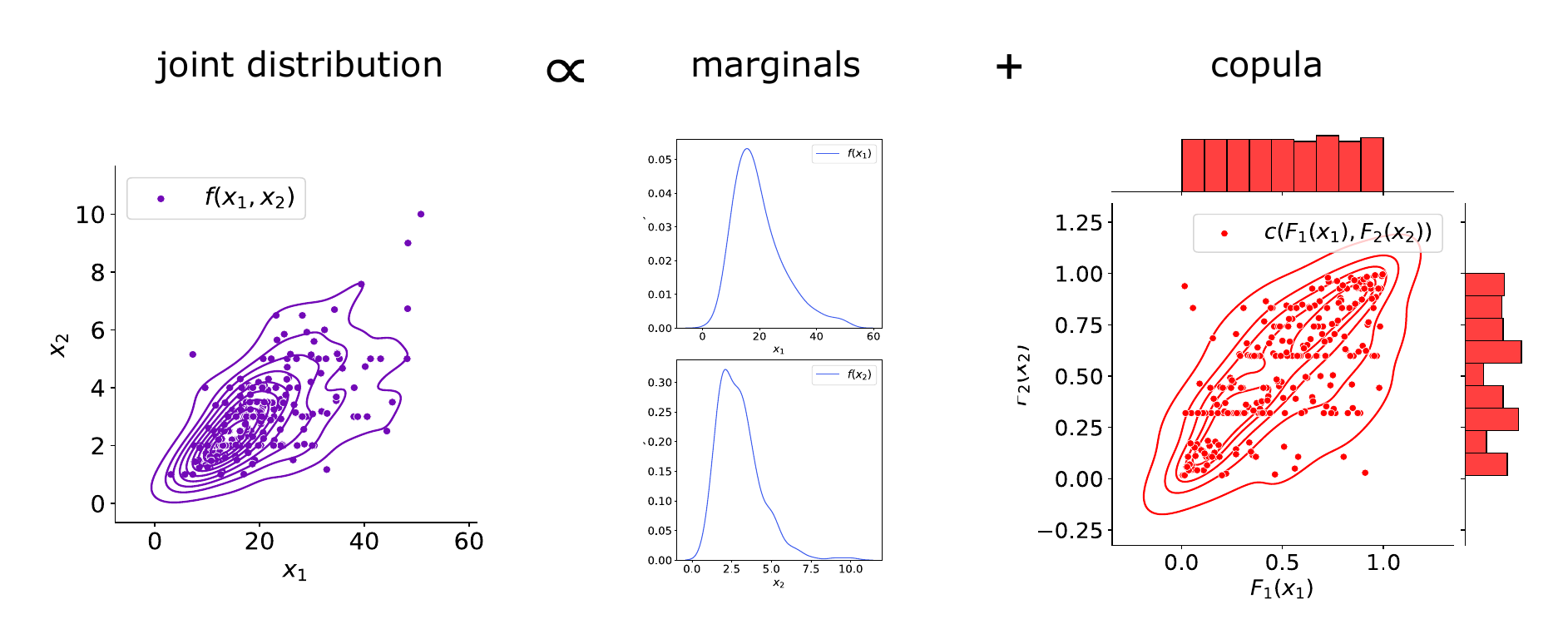}
  \end{center}
  \begin{center}
    \includegraphics[width=0.42\textwidth]{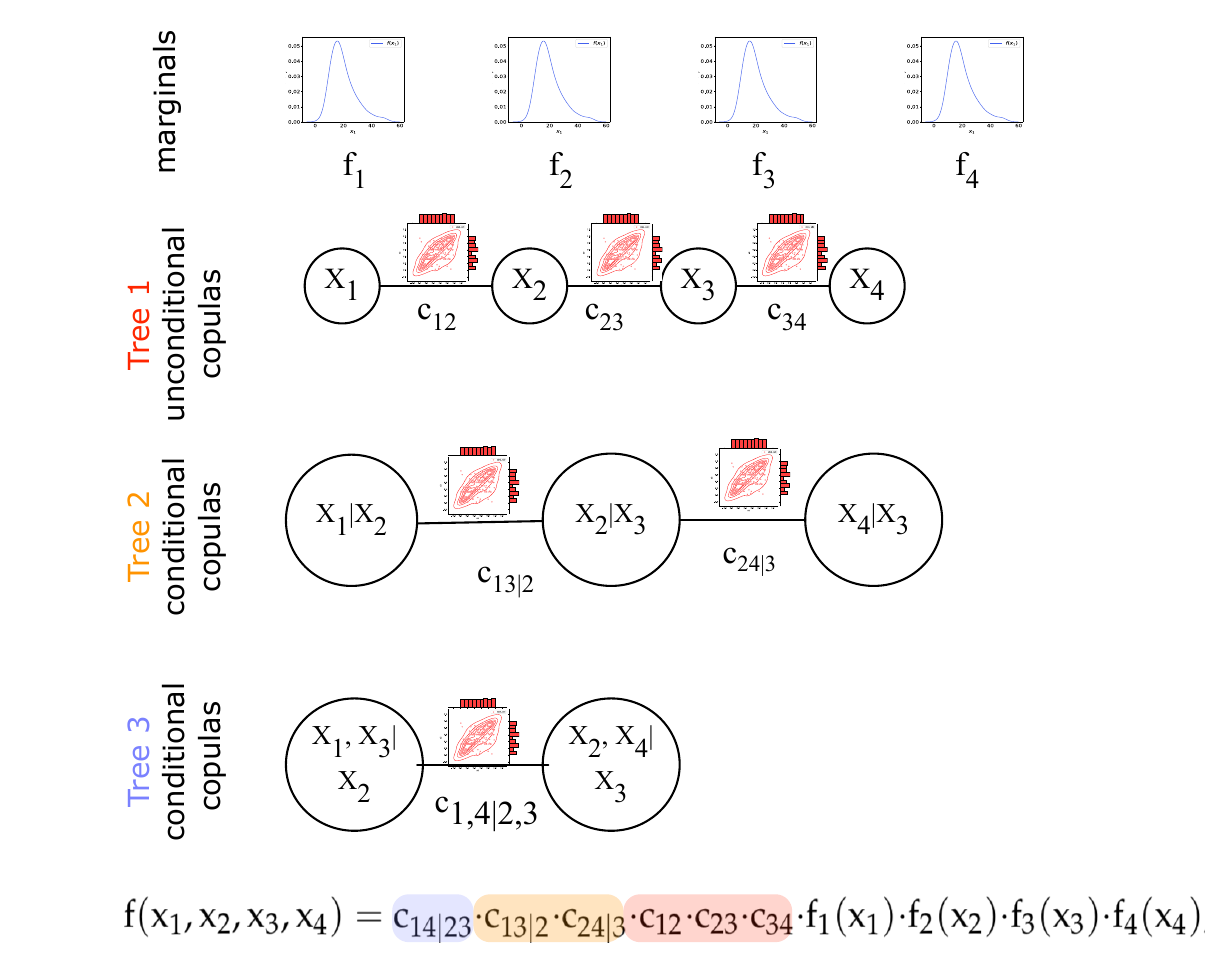}
  \end{center}
  \vspace{-0.5cm}
  \caption{Top: expressing joint densities with copulas; Bottom: Multivariate joint density factorized with a vine copula.}
\end{wrapfigure}\label{fig:vine_example}
\label{fig:copula}
According to \cite{Joe97,Bedford02}, any $M$-dimensional copula density can be decomposed into a product of $\frac{M(M - 1)}{2}$ bivariate (conditional) copula densities.
Although such factorization may not be unique, it can be organized in a graphical model, as a sequence of $M - 1$ nested trees, called \emph{vines}.
We denote a tree as $T_m = (V_m, E_m)$ with $V_m$ and $E_m$ the sets of nodes and edges of tree $m$ for $m = 1, \dots, M-1$. Each edge $e$ is associated with a bivariate copula. An example of a vine copula decomposition is given in \autoref{fig:vine_example}.

In practice, in order to construct a vine, one chooses two components: 
(1) the structure, the set of trees $T_m = (V_m, E_m)$ for $m \in [M-1]$
and (2) the pair copulas, the models for $ c_{j_e, k_e | D_e}$ for $e \in E_m$ and $m \in [M-1]$.

Corresponding algorithms exist for both of those steps and in the rest of the paper, we assume consistency of the vine copula estimators for which we use the implementation by \cite{Nagler2016}, namely its Python version -\verb|pyvinecopulib|.

\subsection{Complexity of the copula estimation}

The complexity for fitting the vine copulas as currently implemented scales as $O(n_{\rm total} M p)$ in the case of density estimation, where $n_{\rm total}$ is the number of points being fit and $p$ is the vine depth. Both estimation and sampling involve a double loop over $M$ and $p$ with an internal step scaling linearly with $n_{\rm total}$. The computational complexity is linearly impacted by $L$ (number of predictive samples). For {BOtied} v1, we have $n_{\rm total} = n$, so this translates to $O(n L M p)$, where $n$ is the number of query candidates, while for BOtied v2, we use the expectation of the posterior samples only, so $n_{\rm total} = nL$ and the complexity remains as $O(n  M p)$. Note that $p \in [M]$ can be truncated for additional efficiency. 

\begin{figure*}
    \centering
    \includegraphics[width=0.95\textwidth]{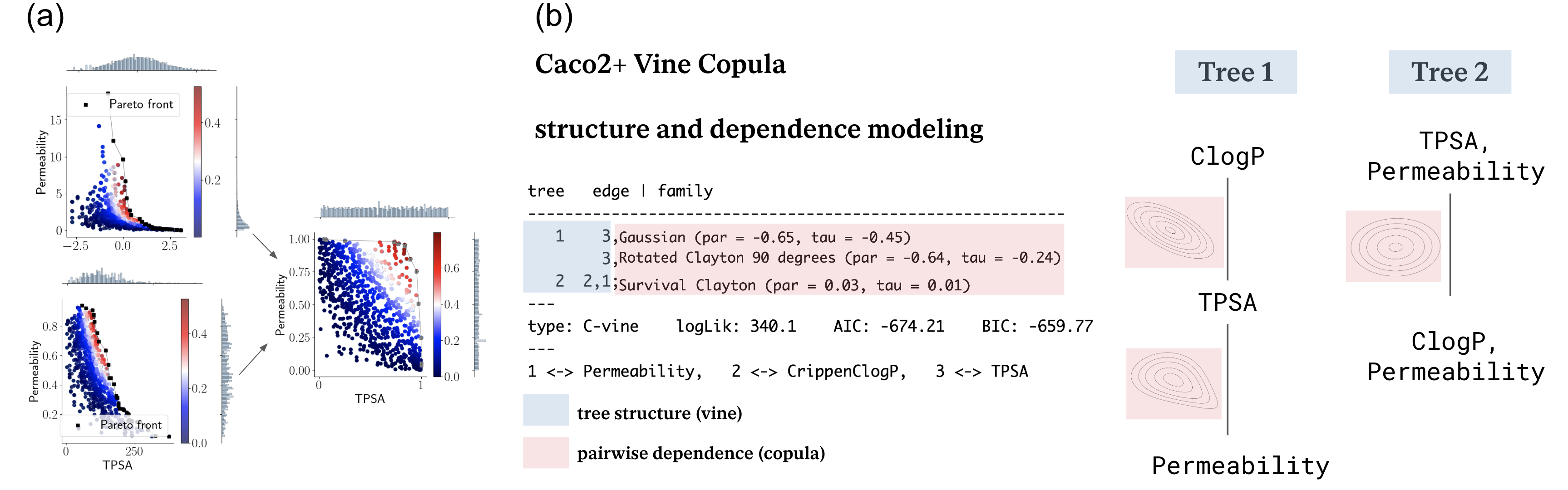}
    \caption{(a). Regardless of the distributions of the marginals, the CDF score from a copula is the same. (b) An example of explicitly encoding domain knowledge in a BO procedure by imposing the \textcolor{blue}{blue} tree structure (specifying the matrix representation of the vine) and \textcolor{pink}{pink} selection of pairwise dependencies (choice of parametric/non-parametric family).}
    \label{fig:caco2_vine}
    \vspace{-0.5cm}
\end{figure*}

\subsection{Copulas in BO}

In the low-data regime, empirical Pareto frontiers tend to be noisy. When we have access to domain knowledge about the objectives, we can use it to construct a model-based Pareto frontier using vine copulas. This section describes how to incorporate (1) the known correlations among the objectives to specify the tree structure (vine) and (2) the pairwise joint distributions (including the tail behavior), approximately estimated from domain knowledge, when specifying the copula models.

The advantages of integrating copula-based estimators for our metric and acquisition function are threefold: (i) scalability from the convenient pair copula construction of vines, (ii) robustness wrt marginal scales and transformations thanks to inherent copula properties \autoref{col1} and \autoref{col2}, and (iii)  domain-aware copula structures from the explicit encoding of dependencies in the vine copula matrix, including choice of dependence type (e.g., low or high tail dependence).

\autoref{fig:caco2_vine} illustrates the use of copulas in the context of optimizing multiple objectives in drug discovery, where data tends to be sparse. In panel (a) we see that, thanks to the separate estimation of marginals and dependence structure, different marginal distributions have the same Pareto front in the PIT space, in which we evaluate our CDF scores. Hence, with copula-based estimators, we can guarantee robustness without any overhead for scalarization or standardization of the data as required by other counterparts. In panel (b) we show how we can encode domain knowledge of the interplay between different molecular properties in the Caco2+ dataset. Namely, permeability is often highly correlated with ClogP and TPSA, with positive and negative correlation, respectively, which is even more notable at the tails of the data (see panel (a) and \autoref{app:experimental_detail}). 
Such dependence can be encoded in the vine copula structure and in the choice of copula family for each pair. For example, we specified a rotated Clayton copula so that the tail dependence between TPSA and permeability is preserved.

\section{Other multivariate CDF estimators}
\label{app:other_cdf_scores}

Copulas are not the only statistical tool we can use for estimating multivariate CDFs. Here we include three more alternatives for the CDF acquisition function based on: empirical CDF, kernel density estimation and multivariate Gaussian. However, not all of them enjoy the fast computation in higher dimensions as vine copulas, and they all lack the guarantees for invariance to scale and transformation.
The sensitivity analysis doesn't show significant difference between the performance of the estimators, thus, the choice can be made based on users' preference. 

We want to highlight the general form of our proposed score, by showing how the CDF estimator as well as the BOTIED acquisition function can be computed with other parametric and non-parametric estimators. In what follows we include: 
\begin{itemize}
    \item \textbf{Multivariate Gaussian CDF} (${{\rm BOtied}}_{\rm mvn}$) We compute the sample mean and covariance $(\mu, \Sigma)$ from the training data, and than use a closed-form analytical solution to obtain the multivariate Gaussian distribution with which we can compute our CDF scores.
    $
        \hat{F}(\mathbf{x}) = P(\mathbf{X}\le \mathbf{x}) \quad where \quad \mathbf{X}\sim(\mathbf{\mu, \Sigma})
    $
    \item \textbf{Empirical CDF} (${{\rm BOtied}}_{\rm empirical}$) The empirical cumulative distribution function is a step function that jumps up by $\frac{1}{n}$ at each of the $n$ data points. Its value is the fraction of observations of the measured variable that are less than or equal to the specified value: 
    $
        \hat{F}_n(t) = \frac{\#elements  \quad in \quad sample \quad < t}{n} = \frac{1}{n}\sum_{i=1}^n \mathbf{1}_{X_i < t}
    $
    \item \textbf{Kernel density estimation} (${{\rm BOtied}}_{\rm KDE}$)
    Finally, we can also use a mixture of density estimators, such as KDE. Since the density is $\hat{f}(x) = \frac{1}{m}\sum_{i=1}^M f_i(x)$, then the joint CDF is the mixture of CDFs, $\hat{F}(x) = \frac{1}{m}\sum_{i=1}^M F_i(x)$. With a Gaussian kernel we have $f_i(x) = \frac{\phi(x-x')}{\sigma}$ and analogously  $F_i = \frac{\Phi(x-x')}{\sigma}$ where $\sigma$ is the kernel bandwidth.
\end{itemize}

\section{Experimental detail and additional results}
\label{app:experimental_detail}

We executed batched BO simulations with sequential greedy optimization and varying batch sizes $q \in \{1, 2, 4\}$. The number of iterations $T$ varied across the experiments.

{\paragraph{Sequential greedy optimization} In batch BO, we seek joint optimization over the $q$ design points, so the decision variable is effectively $q{\rm \times}d$-dimensional. When $q$ is large, we may employ a sequential greedy scheme, where the $q$ designs are selected in series by fantasizing observations at the predictive mean of already-selected designs and conditioning on them to select the next design \citep{wilson2018maximizing}. For the baseline acquisition functions supported in the {BoTorch}, we use the \texttt{optimize\_acqf} function with \texttt{sequential=True}. 

Other parameters include: the initial data size $N_0$, the size of the pool $N$, and the number of predictive posterior samples $L$. We fixed the size of the pool relative to the selected batch, at $N/B=100$. We also fixed $L=20$, which was found to yield good sample coverage and a stable {BOtied} acquisition value.

Unless otherwise stated, the surrogate model was a multi-task Gaussian process (MTGP) with a Matern kernel implemented in \verb|BoTorch| \citep{balandat2020botorch} and \verb|GPyTorch| \citep{gardner2014bayesian}. The inputs and outputs were both scaled to the unit cube for fitting the MTGP, but the outputs were scaled back to their natural units for evaluating the respective acquisition functions. 

\subsection{Branin-Currin} \label{app:bc}
Branin-Currin \citep[$d{\rm =}2$, $M{\rm =}2$][]{belakaria2019max} is a composition of the Branin and Currin functions featuring a concave Pareto front (in the maximization setting). We maximize

\begin{align*}
    f_1(x_1, x_2) = -\left(x_2 - \frac{5.1}{4 \pi^2}x_1^2 + \frac{5}{\pi}x_1 - r \right)^2 + 10(1 - \frac{1}{8 \pi}) \cos(x_1) + 10 \\
    f_2(x_1, x_2) = -[ 1 - \exp \left(-\frac{1}{2 x_2} \right) ] \frac{2300 x_1^3 + 1900 x_1^2 + 2092 x_1 + 60}{100 x_1^3 + 500 x_1^2 + 4 x_1 + 20},
\end{align*}
where $x_1, x_2 \in [0, 1]$. We used $T=30$.

\subsection{DTLZ2}

We took two configurations of DTLZ2 with $d{\rm =}6, M{\rm =}4$ and $d{\rm =}7, M{\rm =}6$ \citep{deb2005searching}.

\subsection{Penicillin production}

The penicillin production problem \citep[$d{\rm =}7$, $M{\rm =}3$;][]{liang2021scalable} simulates the penicillin yield, time to production, and undesired byproduct for seven input parameters of the production process.

\begin{table}
\begin{center}

\caption{HV indicators (computed in the original units) and $I_{\rm CDF}$ different batch size on for the Penicillin dataset. Higher is better and best per column is marked in bold. We report the average metric across twenty random seeds along with their standard error in parentheses.}
\resizebox{0.95\textwidth}{!} 
{ 
\begin{tabular}{lccccccc}
\hline
\multicolumn{1}{c}{} & \multicolumn{2}{c}{\textbf{Penicillin (M=3, q=1)}}& \multicolumn{2}{c}{\textbf{Penicillin (M=3, q=2)}}& \multicolumn{2}{c}{\textbf{Penicillin (M=3, q=4)}}                            \\ \hline
\textbf{}            & \multicolumn{1}{c}{\textbf{$I_{\rm CDF}$}} & \multicolumn{1}{c}{\textbf{HV}} & \multicolumn{1}{c}{\textbf{$I_{\rm CDF}$}} & \multicolumn{1}{c}{\textbf{HV}} & \multicolumn{1}{c}{\textbf{$I_{\rm CDF}$}} & \multicolumn{1}{c}{\textbf{HV}} \\ \hline
\textbf{{BOtied}  v1}       &0.15(0.06) & \textbf{32.69e4(1.78e4)} &\textbf{0.33(0.09)}&\textbf{34.08e4(2.7e4)} &\textbf{0.31(0.08)}&33.55e4(2.4e4)\\
\textbf{{BOtied}  v2}     &\textbf{0.26(0.08)} & 31.05e4(2.11e4)  &0.31(0.07)& 30.06e4(2.21e4)  &0.29(0.07)&\textbf{33.34e4(2.2e4)}\\
\textbf{NParEGO}     &0.19(0.08) & 31.31e4(1.96e4) &0.18(0.06)&30.79e4(2.26e4) &0.18(0.08)&31.67e4(1.4e4)\\
\textbf{NEHVI}       &0.16(0.09) &  30.72e4(2.06e4) &0.19(0.08)&32.04e4(1.85e4) &0.19(0.01)&32.2e4(2.8e4)\\
\textbf{Random}    &0.34(0.05)&10804(112305) &0.21(0.11))&30.8e4(1.86e4) &0.18(0.09)&30.65e4(1.5e4)\\
\hline

\end{tabular}
}
\end{center}
\end{table}

\subsection{Caco2+}
For the Caco2 problem \citep[$M=3$;][]{wang2016adme} the objective is to identify molecules with maximum cell permeability. Here, permeability describes the degree to which a molecule passes through a cellular membrane. This property is critical for drug discovery (DD) programs where the disease protein being targeted resides within the cell (intracellularly). In each experiment, a molecule ${\bm x}_i$ is applied to a monolayer of Caco2 cells and, after incubation, the concentration $c$ of ${\bm x}_i$ is measured on both the input and output side of the monolayer, giving $c_{\rm in}$ and $c_{\rm out}$\citep{van2005caco}. The ratio $c_{
\rm out}/{c_{\rm in}}$ is then treated as the final permeability label $y^p_i$.

Cellular membranes are composed of a complex mixture of lipids and other biomolecules. In order to enter and (passively) diffuse through a membrane, molecule ${\bm x}_i$ should interact favorably with these biomolecules and/or avoid disrupting their packing structure. Increasing the lipophilicity (logP) of ${\bm x}_i$ is thus one strategy to increase permeability. However, increasing logP often results in promiscuous binding of ${\bm x}_i$ to non-disease related proteins, which can lead to undesired side-effects. As such, we seek to minimize the computed logP (clogP, $y^l_i$) in our optimization task and note that this could directly compete with (i.e., harm) permeability.

Lastly and related, common objectives during MPO in DD settings include increasing the affinity and specificity of target binding. As opposed to non-specific lipophilic interactions as above, polar contacts (such as hydrogen bonds) between drug molecules and proteins often result in higher affinity and more specific binding. We compute the topological polar surface area (TPSA, $y^t_i$) of each candidate $\textbf{x}_i$ as one indicator of its ability to form such interactions and seek to maximize it in our optimization. As with decreasing logP, increasing TPSA can negatively impact permeability and we thus consider it a competing objective.

It is important to note that the treatment of each of these optimization tasks as unidirectional (max or min) is a simplification of many practical DD settings. There is often an acceptable range of each value that is targeted, and leaving the bounds in either direction can be problematic for complex reasons. We direct the reader to \cite{d2012multi} for a comprehensive review.

For fitting the MTGP on the Caco2+ data, we represent each input molecule as a concatenation of fingerprint and fragment feature vectors, known as fragprints \citep{thawani2020photoswitch} and use the Tanimoto kernel implemented in \verb|GAUCHE| \citep{griffiths2022gauche}. 

\subsection{Real-world datasets for data-driven evolutionary multi-objective optimization}
\label{app:ddmop}
To evaluate BOtied in real-world scenarios, we include experiments over three datasets from the DDMOP benchmark \citep{ddmop}. Differently from the synthetic test functions which have analytical solutions, DDMOP, proposes a testbed of complex objective functions, approximated by expensive numerical simulations, formulated as Data-Driven Multi-objective Optimization Problems, hence the name DDMOP. We select the three scenarios \footnote{Which have more than 100 data points in the latest version of that datasets we were provided by the authors at the time of writing this paper}:
\begin{itemize}
    \item \textbf{Car cab} from \citep{deb2009reliability}, optimization of vehicle frontal structure, $d=11, M=9, N=120$. The objectives represent the performance of the car cab, through weight of the car, fuel economy, acceleration time, road noise at different speed, and roominess of the car.
    \item \textbf{Power system} \citep{kavasseri2011joint},  $d=11, M=3, N=120$. The objectives relate to the performance of a power system, active power loss, voltage deviation and generation cost based on the optimal joint placement of phasor measurement units.
    \item \textbf{Neural network performance} \cite{jin2008pareto}, $d=17, M=2, N=186$. One objective denotes the complexity of the network in terms of nonzero weights, while the second objective is the classification error rate of the neural network.
\end{itemize}

We negate all three problems to turn them into maximization objectives. As we approach these problems from a realistic perspective, we ran the experiments with batch size $q=4$, $T=20$ and initial set of $24, 9, 10$ points respectively. The reference points for each dataset were chosen as the minimum per objective decreased by 1e-3.

\begin{table}[h]
\begin{center}
\caption{HV and $I_{\rm CDF}$ for three DDMOP datasets. We report the average metric across twenty random seeds along with their standard error in parentheses.}
\vspace{0.1cm}
\resizebox{0.95\textwidth}{!} 
{ 
\begin{tabular}{lcccccc}
\hline
\hline
 & \multicolumn{2}{l}{\textbf{CarCab (M=9)}} & \multicolumn{2}{l}{\textbf{PowerSystem (M=3)}} & \multicolumn{2}{l}{\textbf{NeuralNetwork (M=2)}} \\ \hline
                   & \textbf{$I_{\rm CDF}$} & \textbf{HV} & \textbf{$I_{\rm CDF}$} & \textbf{HV}   & \textbf{$I_{\rm CDF}$} & \textbf{HV} \\  \hline
\textbf{BOtied v1} & \textbf{0.095(0.01)}  & 2.39e5(0.21e5) & \textbf{0.732(002)}   & \textbf{0.0235(0.001)} & \textbf{0.852(0.01)}  & \textbf{0.512(0.02)} \\
\textbf{BOtied v2} & 0.091(0.02)  & \textbf{2.44e5(0.30e5)}  & \textbf{0.732(002)}   & \textbf{0.0235(0.001)} & 0.849(0.02)  & 0.511(0.02) \\
\textbf{NPareGo}   & 0.094(0.02)  & 2.36e5(0.23e5)  & 0.729(0.02)  & 0.0233(0.001) & 0.847(002)   & 0.509(0.03) \\
\textbf{NEHVI}     & 0.090(0.02)  & 2.3e5(0.35e5)  & 0.724(0.02)  & 0.0218(0.002) & 0.847(0.02)  & 0.502(0.03) \\
 \textbf{random}    & 0.079(0.02)  & 2.24e5(0.31e5)  & 0.721(0.05)  & 0.0222(0.002) & 0.850(0.02)  & 0.504(0.03) \\ \hline
\end{tabular}
}
\end{center}
\end{table}

\subsection{Details on wall clock time}

\textbf{Details for \autoref{fig:time}}. For all acquisition functions, we report the wall clock time per single acquisition function evaluation as computed on a Tesla V100 SXM2 GPU (16GB
RAM) and an Intel Xeon CPU @ 2.30GHz (240GB RAM). A single call takes in the surrogate inference results for the candidate pool as well as the previously evaluated points and computes the acquisition scores. 
\begin{itemize}\itemsep1em 
    \item  BC $M{\rm =}2$: $q$ batch size = 4, number of predictive samples=40, initial $n$ = 10, pool size = 40
    \item DTLZ $M{\rm =}4$: $q$ batch size = 4, number of predictive samples=20, initial $n$ = 50, pool size = 40
    \item DTLZ $M{\rm =}6$: $q$ batch size = 4, number of predictive samples=20, initial $n$ = 50, pool size = 40.
\end{itemize}

\begin{figure}[h!]
    \begin{center}
    \begin{subfigure}[Desirable properties]{\includegraphics[width=0.47\textwidth]{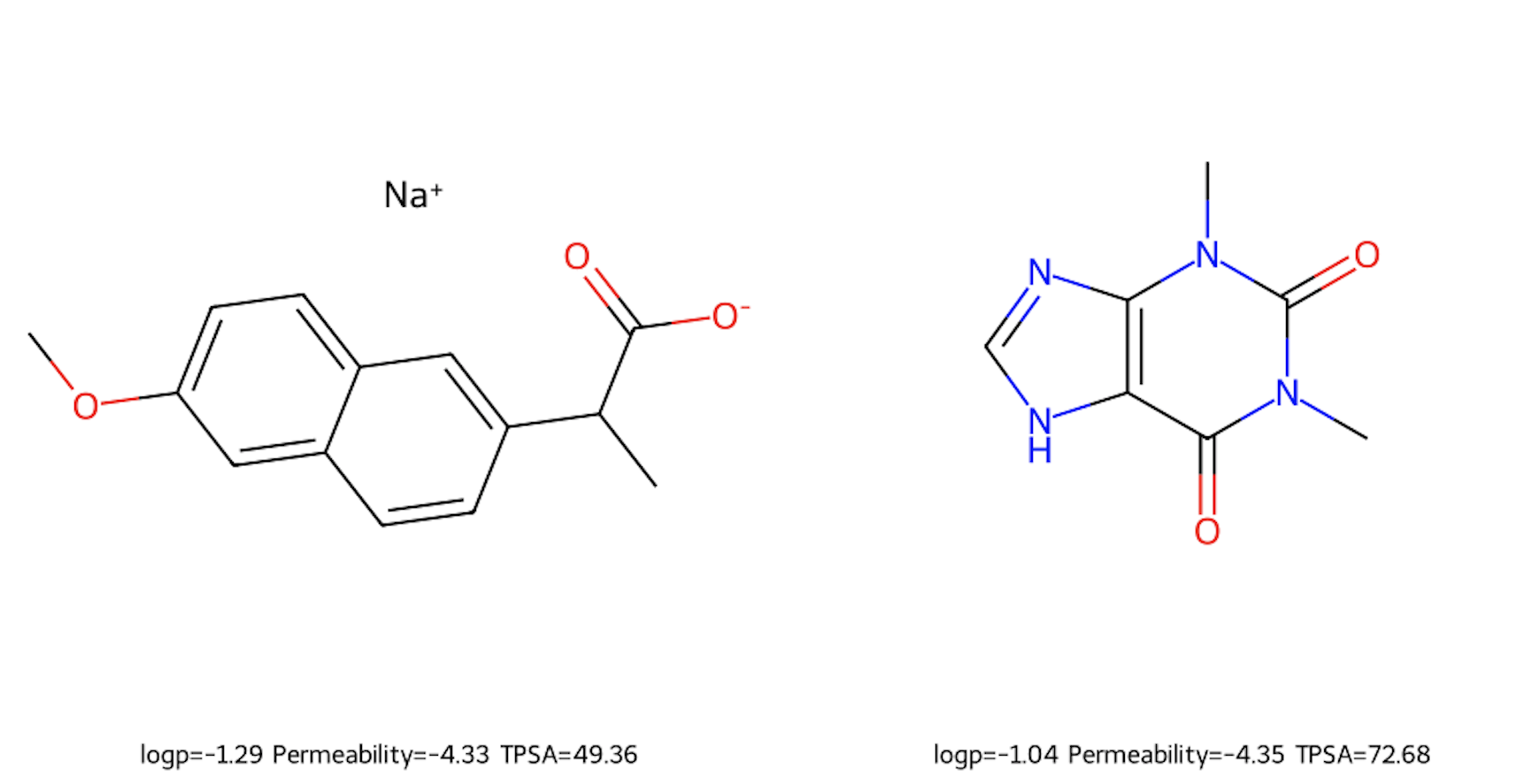}}
    \label{fig:mol_good}
    \end{subfigure}
    \begin{subfigure}[Undesirable properties (high log p, low permeability)]{
    \includegraphics[width=0.31\textwidth]{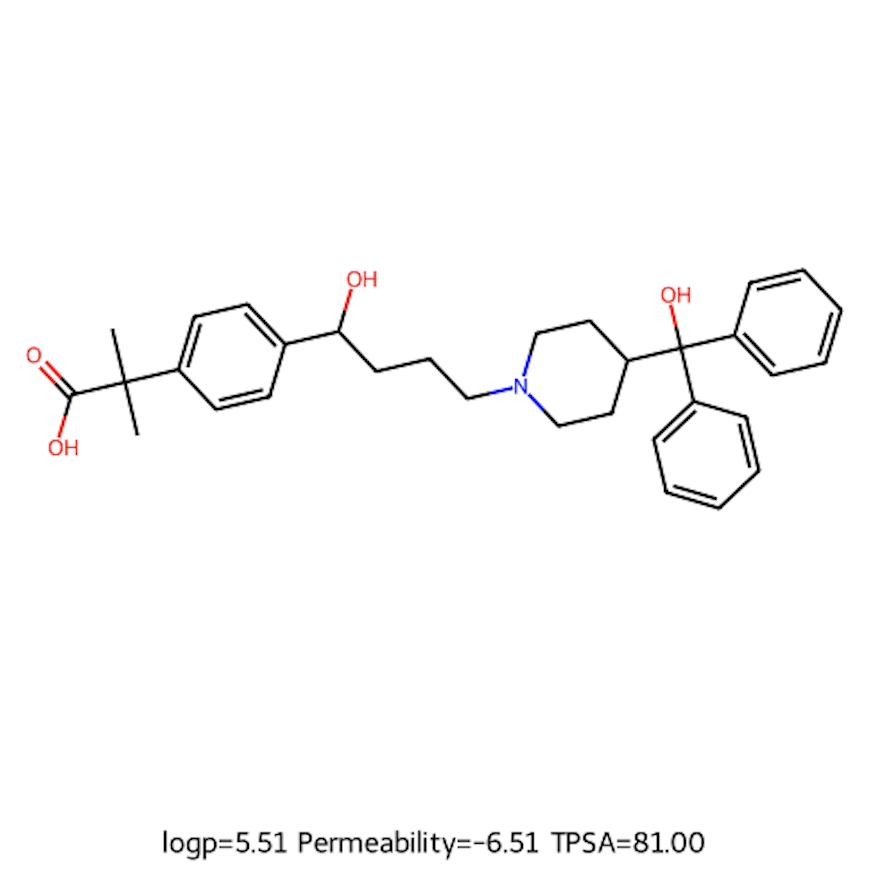}}
        \label{fig:mol_bad}
        \end{subfigure}
    \end{center}
    \caption{Examples of molecules in the Caco2+ dataset. The goal for the Caco2+ problem is to minimize log p, maximize permeability, and maximize TPSA.}
\end{figure}

\section{Ablation studies}
\label{app:ablation}

\begin{figure}[H]
\begin{center}
    \begin{subfigure}[Varying the number of posterior samples]
        {\includegraphics[width=0.45\textwidth]{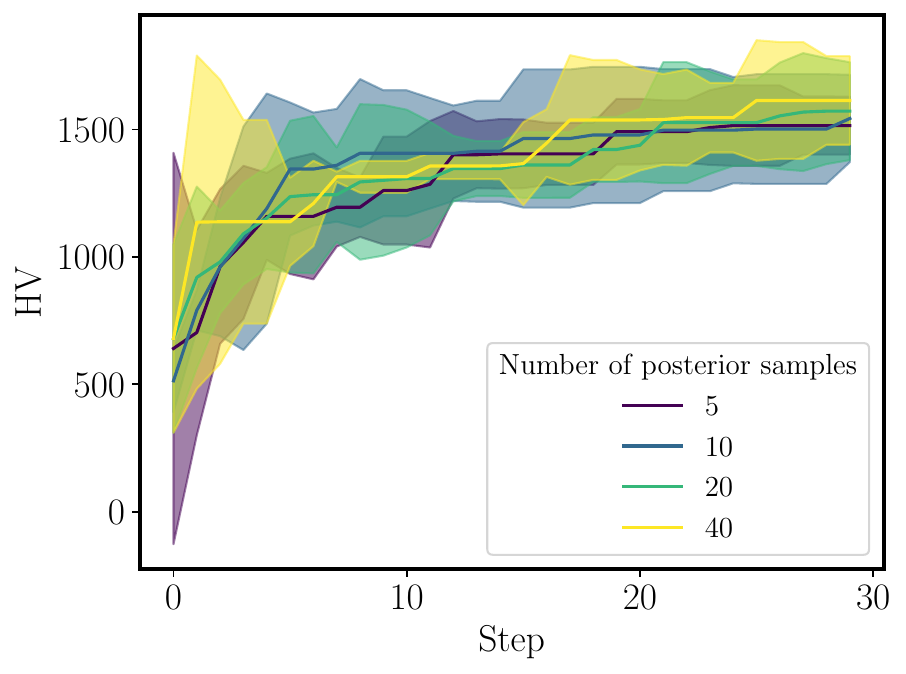}}
        \label{fig:posterior_samples}
    \end{subfigure}
    \begin{subfigure}[Varying the batch size]
        {\includegraphics[width=0.45\textwidth]{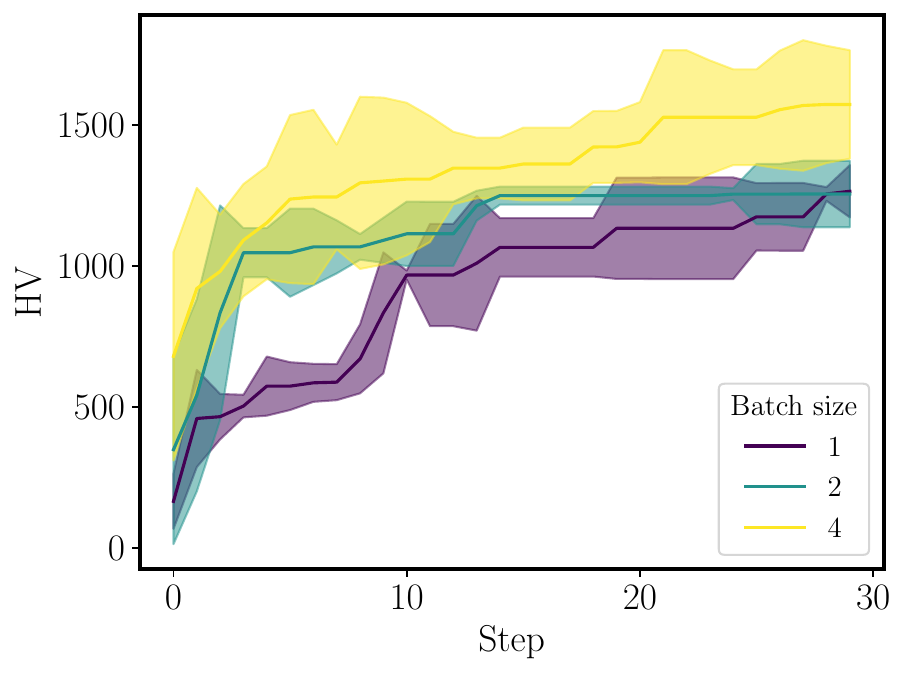}}  
        \label{fig:batch_size}
    \end{subfigure}
\end{center}
\caption{Ablation studies for BOtied v1. (a) BOtied is robust to the number of posterior samples drawn. (b) Increasing the batch size improves acquisition, particularly as it improves the CDF fit quality in earlier iterations.}
\end{figure}

\section{Importance of invariance to scaling and monotonic transformations} \label{app:invariance_examples}

Consider a scenario that occurs commonly in drug design, where both objectives are ``zero-inflated,'' meaning that they are distributed with an abundance of zero (null) values plus a wide dispersion of valid, non-null values (\autoref{fig:scenario_2}). We linearly scale the objective values to the $[0, 1]$ range and define the ``null'' value at 1 for both objectives. The color gradient corresponds to the indicator value at each solution ($q=1$). With HV, we need to specify a reference point, set at [1.1, 1.1] in this case. Because the having a null value in even one of the objectives makes the HV small for a solution, the HV indicator can only distinguish points with non-null values in 
\textit{both} objectives (lower left corner) from all other points. It assigns near-zero scores to regions with null values in only one objective (upper left and lower right corners), which should be included in the approximate Pareto front. On the other hand, the CDF indicator effectively identifies the full Pareto front, including the upper left and lower right corners.

\begin{figure}
    \centering
    \begin{subfigure}[HV indicator]
        {\includegraphics[trim={0.6cm 0cm 0 0cm}, clip, width=0.45\textwidth]{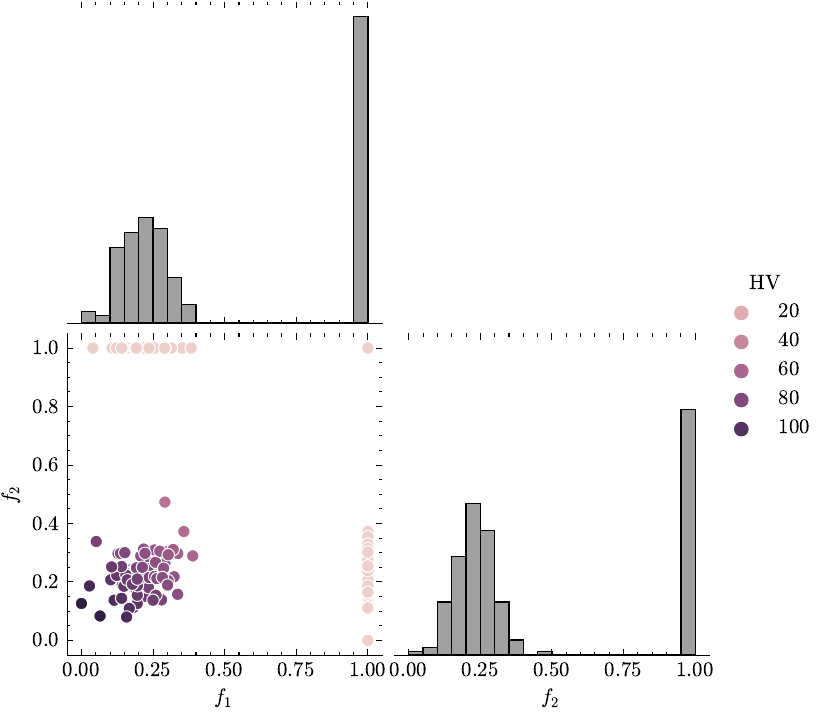}}	
        \end{subfigure}
    \begin{subfigure}[CDF indicator]
        {\includegraphics[clip, width=0.45\textwidth]{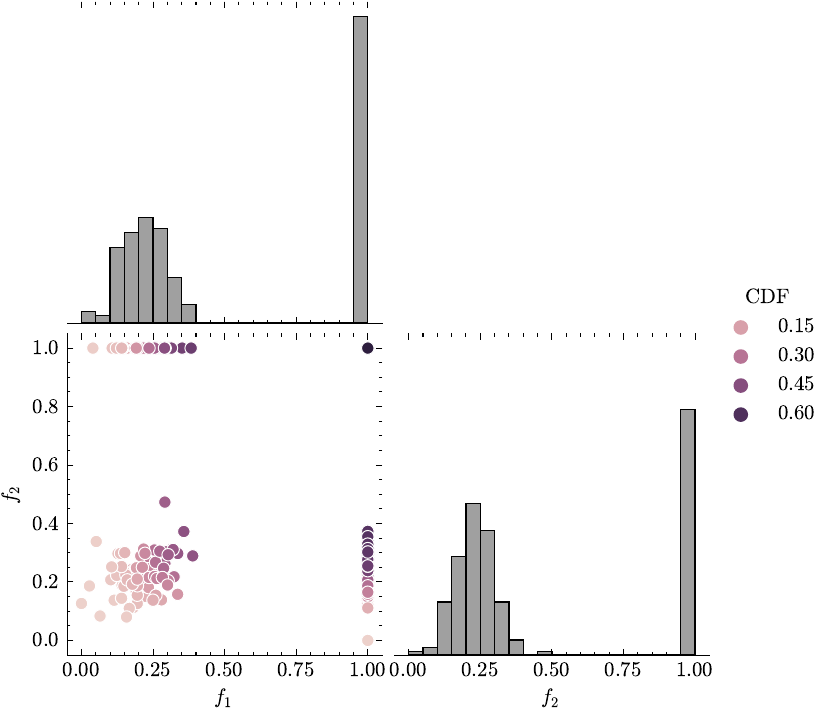}}	
    \end{subfigure}
     \caption{A scenario where both objectives are "null-inflated," meaning that they are distributed with an abundance of null values plus a wide dispersion of valid, non-null values. We linearly scale the objective values to the $[0, 1]$ range and define the "null" value at 1 for both objectives. The color gradient corresponds to the indicator value at each solution ($q{\rm =}1$). (a) With HV, we need to specify a reference point, set at [1.1, 1.1] in this case. Because the having a null value in even one of the objectives makes the HV small for a solution, the HV indicator can only distinguish points with non-null values in {\it both} objectives (lower left corner) from all other points. It assigns near-zero scores to regions with null values in only one objective (upper left and lower right corners), which should be included in the approximate Pareto front. (b) The CDF indicator effectively identifies the full Pareto front, including the upper left and lower right corners.}
     \label{fig:scenario_2}
 \end{figure}
 
\section{Discontinuous Pareto fronts} \label{app:copula_mixture}

Copulas can flexibly model multi-modal outcome distributions as well, particularly those with discontinuous Pareto fronts. In Figure \ref{fig:discont_pf}, we consider objectives distributed as a mixture of two well-separated Gaussians. On the 200 simulated observations, we fit a CDF with a Gaussian mixture copula and kernel density estimation (KDE) marginals. The zero level line of the CDF closely traces the true Pareto front (solid red curve).

\begin{figure}[h]
    \centering
    \includegraphics[width=0.6\textwidth]{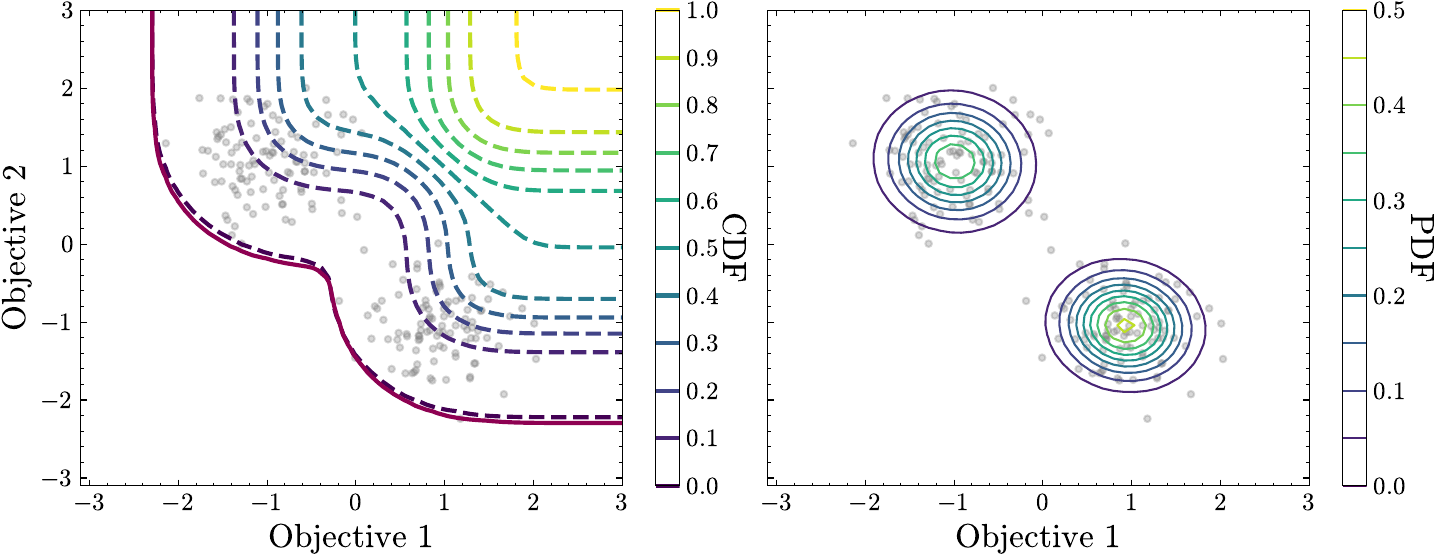}
     \caption{Level lines of the CDF (left) and the PDF (right) from a CDF fit with Gaussian mixture copula and kernel density estimation (KDE) marginals, based on 200 observations simulated from a mixture of two Gaussians (gray dots). The zero level line of the CDF closely traces the true Pareto front (solid red curve).}
     \label{fig:discont_pf}
 \end{figure}


\end{document}

%% file: math_commands.tex
\usepackage{amsmath,amssymb,amsthm,enumitem,amsfonts,nccmath,bm}
\usepackage{mathrsfs}
\usepackage{mathtools}

\def\eqref#1{equation~\ref{#1}}

\def\1{\bm{1}}

\DeclareMathAlphabet{\mathsfit}{\encodingdefault}{\sfdefault}{m}{sl}
\SetMathAlphabet{\mathsfit}{bold}{\encodingdefault}{\sfdefault}{bx}{n}

\DeclareMathOperator*{\argmax}{arg\,max}